%% file: main.tex
\documentclass{applemlr}

\input{apple_preamble}

\usepackage{nicefrac}
\usepackage[normalem]{ulem}

\definecolor{paramblue}{RGB}{30,100,200}
\newcommand{\pp}[1]{{\color{paramblue}#1}}

\setlist[itemize]{noitemsep, topsep=0pt}

\title{Scaling Laws for Mixture Pretraining Under Data Constraints}

\author[1,2]{Anastasiia Sedova}
\author[1]{Skyler Seto}
\author[1]{Natalie Schluter}
\author[1]{Pierre Ablin}

\affiliation[1]{Apple}
\affiliation[2]{ITU}

\abstract{%
As language models scale, the amount of data they require grows -- yet many target data sources, such as low-resource languages or specialized domains, are inherently limited in size.
A common strategy is to mix this scarce but valuable target data with abundant generic data, which presents a fundamental trade-off: too little target data in the mixture underexposes the model to the target domain, while too much target data repeats the same examples excessively, yielding diminishing returns and eventual overfitting.
We study this trade-off across more than 2,000 language-model training runs spanning multiple model and target dataset sizes, as well as several data types, including multilingual, domain-specific, and quality-filtered mixtures.
Across all settings, we find that repetition is a central driver of target-domain performance, and that mixture training tolerates much higher repetition than single-source training:
scarce target corpora can be reused 15--20 times, with the optimal number of repetitions depending on the target data size, compute budget, and model scale.
Next, we introduce a repetition-aware mixture scaling law that accounts for the decreasing value of repeated target tokens and the regularizing role of generic data.
Optimizing the scaling law provides a principled way to compute effective mixture configurations, yielding practical mixture recommendations for pretraining under data constraints.%
}

\metadata[Correspondence]{\sffamily Anastasiia Sedova: \url{asedova@apple.com}
}
\date{\sffamily\today}

\begin{document}

\maketitle

\input{chapters/01_intro}
\input{chapters/02_rw}

\input{chapters/03_experimental_setup}

\input{chapters/04_results}

\input{chapters/05_scaling_law}

\input{chapters/06_ablations}

\input{chapters/07_conclusion}

\bibliographystyle{plainnat}
\bibliography{references}

\beginappendix
\input{chapters/09_appendix}

\applefootnote{ \textcolor{textgray}{\sffamily Apple and the Apple logo are trademarks of Apple Inc., registered in the U.S. and other countries and regions.}}

\end{document}

%% file: apple_preamble.tex
\usepackage{amsmath}
\usepackage{enumerate}
\usepackage{algorithm}
\usepackage{algpseudocode}
\usepackage{amsfonts}
\usepackage{amsthm}
\usepackage{diagbox}
\usepackage{colortbl}
\usepackage{amssymb}
\usepackage{xspace}
\usepackage{wrapfig}
\usepackage{adjustbox}
\usepackage{tabularx}
\usepackage{mathtools}
\usepackage{tikz}
\usepackage{enumitem}
\usepackage{silence}
\usepackage{dsfont}
\usepackage{makecell}
\usepackage{xfakebold}
\input{math_commands}

\definecolor{textgray}{HTML}{6E6E73}
\usetikzlibrary{positioning, calc}
\usetikzlibrary{decorations.pathmorphing}

\makeatletter
\patchcmd{\wrong@fontshape}{\@gobbletwo}{}{}{}
\makeatother
\numberwithin{equation}{section}
\makeatletter
\AtBeginDocument{
  \urlstyle{sf}
  
}
\makeatother

\definecolor{light}{RGB}{125, 125, 125}
\crefname{tcb@cnt@pbox}{code}{code}
\Crefname{tcb@cnt@pbox}{Code}{Code}
\crefname{assumption}{assumption}{assumption}
\Crefname{assumption}{Assumption}{Assumptions}

\newtcolorbox[auto counter]{pbox}[2][]{
  colback=white,
  title=Code~\thetcbcounter: #2,
  #1,fonttitle=\sffamily,
  fontupper=\sffamily,
  arc=2pt,
  colframe=bgcolor,
  coltitle=fgcolor,
  colbacktitle=bgcolor,
  toptitle=0.25cm,
  bottomtitle=0.125cm
}

\makeatletter
\newcommand\applefootnote[1]{%
  \begingroup
  \renewcommand\thefootnote{}%
  \renewcommand\@makefntext[1]{\noindent##1}%
  \footnote{#1}%
  \addtocounter{footnote}{-1}%
  \endgroup
}
\makeatother

\definecolor{cverbbg}{gray}{0.90}

%% file: math_commands.tex
\usepackage{amsmath,amsfonts,bm}

\def\eqref#1{equation~\ref{#1}}

\def\1{\bm{1}}

\DeclareMathAlphabet{\mathsfit}{\encodingdefault}{\sfdefault}{m}{sl}
\SetMathAlphabet{\mathsfit}{bold}{\encodingdefault}{\sfdefault}{bx}{n}

%% file: chapters/01_intro.tex
\begin{figure*}[h!]
\centering
\vspace{-10pt}
\includegraphics[width=0.82\textwidth]{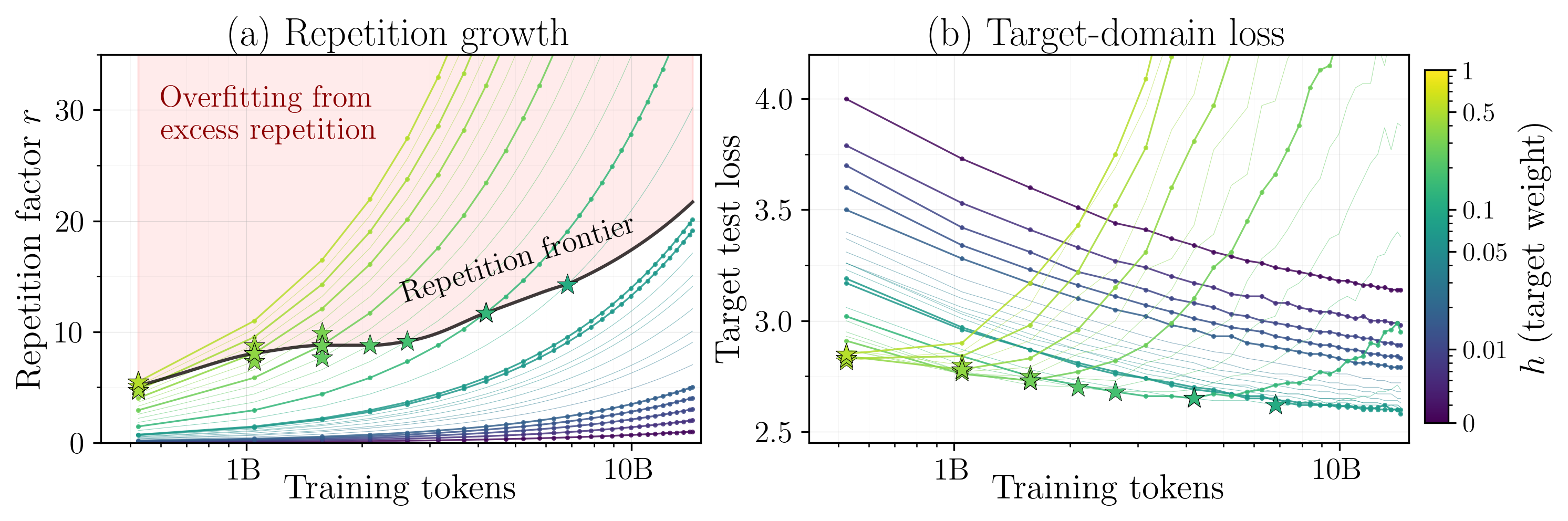}
\vspace{-5pt}
\caption{Repetition dynamics for 143M model, 50M German tokens mixed with English, $h$ defining the weight of German data in the training mix.
\textbf{(a)}~Repetition factor $r$ grows with training tokens; beyond the repetition frontier, loss begins to increase. \textbf{(b)}~German validation loss vs.\ training tokens. Stars indicate overfitting onset.}
\vspace{-10pt}
\label{fig:main}
\end{figure*}
\section{Introduction}
Large language models (LLMs) have demonstrated remarkable performance across a wide range of language understanding, mathematics, science, and knowledge-intensive tasks \citep{luong2025advanced,woodruff2026accelerating,singh2025openai,antropic2026}. 
While much of this success can be attributed to large scale pretraining corpora exceeding trillions of tokens \citep{hojel2025essential,li2024datacomp}, many language model pretraining scenarios involve data that cannot be freely scaled including low-resource languages, specialized domains, and curated datasets, which offer far less unique data.
During pretraining, this data is mixed with abundant generic data, such as pairing low-resource language text with English text, or a domain-specific math corpus with generic web text.

\begin{wrapfigure}{r}{0.4\textwidth}
\centering
\includegraphics[width=0.4\textwidth]{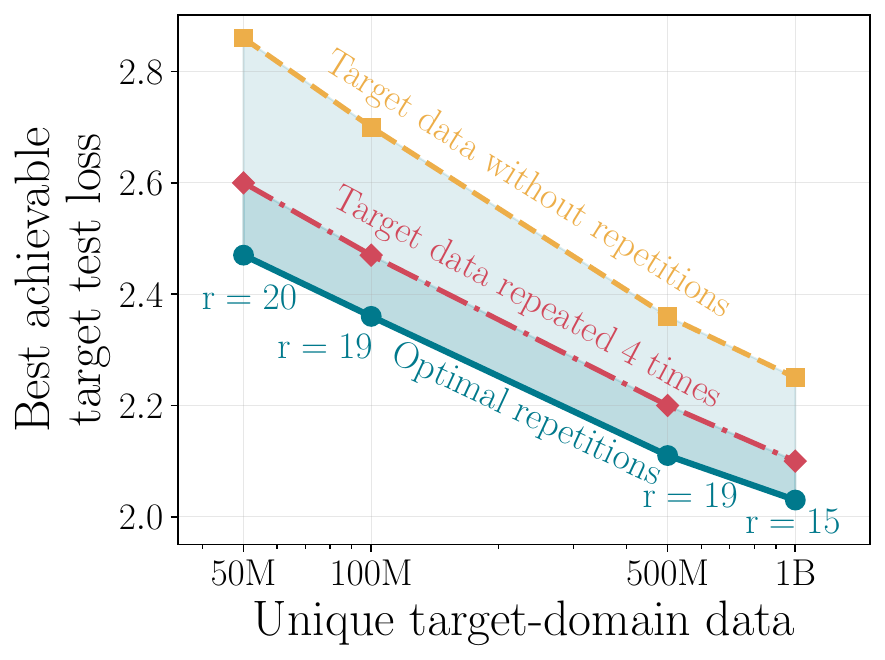}
\vspace{-15pt}
\caption{Best achievable German test loss by target data size, shown for the 539M model.}
\label{fig:diminishing_539M}
\vspace{-5pt}
\end{wrapfigure}

However, mixing introduces a new trade-off. The limited data must be repeated for enough of the total training to provide sufficient target-domain signal, but high repetition leads to memorization and eventual overfitting. Figure~\ref{fig:main} illustrates this challenge for a 143M model with 50M German tokens mixed with English. Higher fractions of German data start overfitting (Figure~\ref{fig:main}b) as the number of repetitions crosses a frontier (Figure~\ref{fig:main}a).
This raises the question: \textit{how do model scale, data size, and repetition jointly shape the outcome of mixture pretraining when target data in the mixture is constrained?}

Prior work has studied these two dimensions separately. 
\citet{muennighoff2023scaling} derive scaling laws for data-constrained training, showing that up to 4 repetitions over a fixed monolithic dataset are nearly as valuable as unique data.
However, this result, widely adopted as a practical ceiling for data repetition, applies to single-source training where \emph{all} tokens are repeated.
In mixture pretraining, only the constrained domain is repeated while the generic domain continues to supply fresh tokens, which is a structurally different regime.
Conversely, data mixing laws \citep{ye2024data, xie2024doremi, shukor2025scalinglawsoptimaldata} optimize the composition of training mixtures to maximize downstream performance, but assume each data source has unlimited data.
Our work sits at the intersection: we study how to optimally mix data-constrained sources with abundant ones, jointly optimizing mixture weights and repetition, a setting that arises in many real-world pretraining scenarios.

Our contributions are:
\begin{itemize}[leftmargin=1.5em]
    \item A systematic empirical study of over 2000 training runs spanning model sizes from 101M to 805M parameters, different target data sizes, and diverse data types: multilingual (German, French, Swahili), multi-domain (mathematics, scientific papers, Wikipedia), and quality-filtered subsets.
    \item Empirical findings on repetition in data mixtures: repeated tokens follow diminishing returns predictably across all data types; optimal repetition scales with target dataset size and compute budget; and larger models consistently extract more from limited data despite overfitting faster. 
    Crucially, abundant generic data sustains learning and unlocks far higher repetition than
    reported for single-source training without performance degradation, with optimal for target task performance repetitions reaching 15-20 times (Figure~\ref{fig:diminishing_539M}).
    \item A scaling law at the intersection of data-constrained training and mixture optimization that predicts target-domain loss as a function of target data size, mixture ratio, and model size. We demonstrate that this scaling law can be fitted at small scales and then extrapolates at larger scales, and that it can be used to estimate accurately the optimal number of repetitions.
\end{itemize}

We also extend our findings to the case of two scarce target domains.
Together, our findings provide both empirical understanding of how repetition behaves in data mixtures and a predictive scaling law for training effectively on limited target data.

%% file: chapters/02_rw.tex
\section{Related Work}

\paragraph{Neural scaling laws}
The study of scaling laws for language models was popularized by \citet{kaplan2020scaling}, who showed that loss decreases as a power law in model size, dataset size, and compute. \citet{hoffmann2022training} refined these findings with the Chinchilla scaling laws, establishing compute-optimal training recipes that balance model size and data. Relating to our work, \citet{muennighoff2023scaling} generalize Chinchilla to a single data-constrained domain, finding that repeating data up to approximately four epochs causes negligible degradation, with meaningful gains extending to roughly 16 epochs and ending at 40. They propose scaling laws that model the diminishing value of repeated tokens. Subsequent work has extended scaling laws to downstream task performance~\citep{wei2022emergent}, mixture-of-experts architectures~\citep{abnar2025parameters,krajewski2024scaling,wang2024scaling}, continued pretraining~\citep{que2024d,liew2025acceleration,seto2026optimal}, finetuning~\citep{bethune2025scaling,zhang2024scaling}, and data quality~\citep{chang2024scaling}.
Our work extends this line of research providing a predictive scaling law for data constrained mixtures, an under-explored regime where only one component of a data mixture is unlimited while others are severely constrained.

\paragraph{Data mixing for language models}
Training on a mixture of data domains has become standard, with corpora aggregated from multiple web sources~\citep{gao2020pile,cerebras2023slimpajama,soldaini2024dolma} with varying data size per domain, and recent models are trained on carefully tuned domain mixtures~\citep{bakouch2025smollm3,olmo2025olmo}. Prior work also focus on optimizing domain weights using distributionally robust optimization  \citep{xie2024doremi,fan2024doge}, or by discovering domains through clustering the pretraining corpus~\citep{diaonemotron,grangier2025task}.
The domain mixture scaling law (DMSL) framework~\citep{ye2024data,shukor2025scalinglawsoptimaldata} models how per-domain loss depends on mixture weights and total compute. Our repetition-aware scaling law builds on DMSL by incorporating the effect of data repetition, which becomes critical when any mixture component is data-limited. Finally, \citet{anonymous2026mixdonttune} compare data mixing against hyperparameter tuning (weight decay, learning rate) as competing strategies for data-constrained pretraining.
The focus of our work is on how to set the mixture optimally once mixing is chosen by characterizing how the mixing ratio, repetition, and model size jointly determine target-domain loss, and that eliminates the need for per-configuration sweeps.

\paragraph{Data-constrained pretraining}
Several works study data-constrained pretraining. \citet{hernandez2022scaling} propose parametric models for the value of repeated data in multi-epoch training, and \citet{goyal2024scaling} compare repeating high-quality data against training on fresh lower-quality data, concluding that the benefit of filtering depends on the total compute budget. When high-quality data for the target domain is constrained, synthetic data has been used as an alternative to repetition. \citet{gunasekar2023textbooks} pretrained on synthetic ``textbook quality'' data, and \citet{maini2024rephrasing} demonstrate that rephrasing web data offer alternatives to repeating data. For multi-domain settings, \citet{seto2025training} study pretraining bilingual models when target-language data is constrained, showing that high-quality auxiliary-language data can partially substitute for target-language data in typologically close languages, but that model scaling has diminishing returns. Low-resource language modeling faces similar constraints, where mixing with high-resource languages is a common strategy ~\citep{joshi2020state}. This work studies different scenarios where data is constrained in at least one target domain, and unconstrained in other generic data, and provides a recipe for determining the target domain loss.  This is tangential to other approaches which aim to improve the diversity of data by building more target data, or incorporating more generic data.

%% file: chapters/03_experimental_setup.tex
\section{Methodology: Mixture Training and Datasets}
\label{sec:datasets}

We consider a mixture pretraining setup with two data sources, where one \textit{target} source has limited size and a \textit{generic} source provides effectively unlimited data. Let $D_\text{target}$ the target data size, i.e., the number of unique target-domain tokens, $D_\text{total}$ the total number of training tokens (i.e., the total compute budget), $h\in[0,1]$ the fraction of $D_\text{total}$ devoted to the target domain.  The number of repetitions $r$ is the number of time the unique target tokens are repeated throughout training:
\begin{equation}\label{eq:main}
r =  h \cdot D_\text{total}/D_\text{target} \enspace.
\end{equation}
For a fixed compute budget $D_\text{total}$ and data pool $D_\text{target}$, increasing the target weight $h$ increases the repetition factor $r$, and vice versa. 
The generic domain is assumed abundant enough to never repeat.
We extend the framework to multiple constrained target domains in Section~\ref{sec:multi_domain}.

To test whether the repetition-diversity trade-off generalizes, we study three types of data-constrained scenarios commonly encountered in practice: limited target-language data (multilingual), limited specialised-domain data (multi-domain), and limited high-quality data (quality filtering).

\paragraph{Multilingual}
The target domain is German text from FineWeb2 \citep{penedo2025fineweb2}, artificially constrained to 50M, 100M, 500M, or 1B tokens\footnote{Across all our experimental setups, each smaller set is a subset of a larger one.}, and an unlimited (no constraint) setting. The generic domain is English web text from FineWeb \citep{penedo2024fineweb}, treated as effectively unlimited. To verify that findings are not language-specific, we run 50M, 100M, and 500M data size experiments with French and Swahili as target languages, also derived from FineWeb2.
Evaluation is done on a held-out set of FineWeb2 for the corresponding language.

\begin{figure}[!t]
\centering
\includegraphics[width=\textwidth]{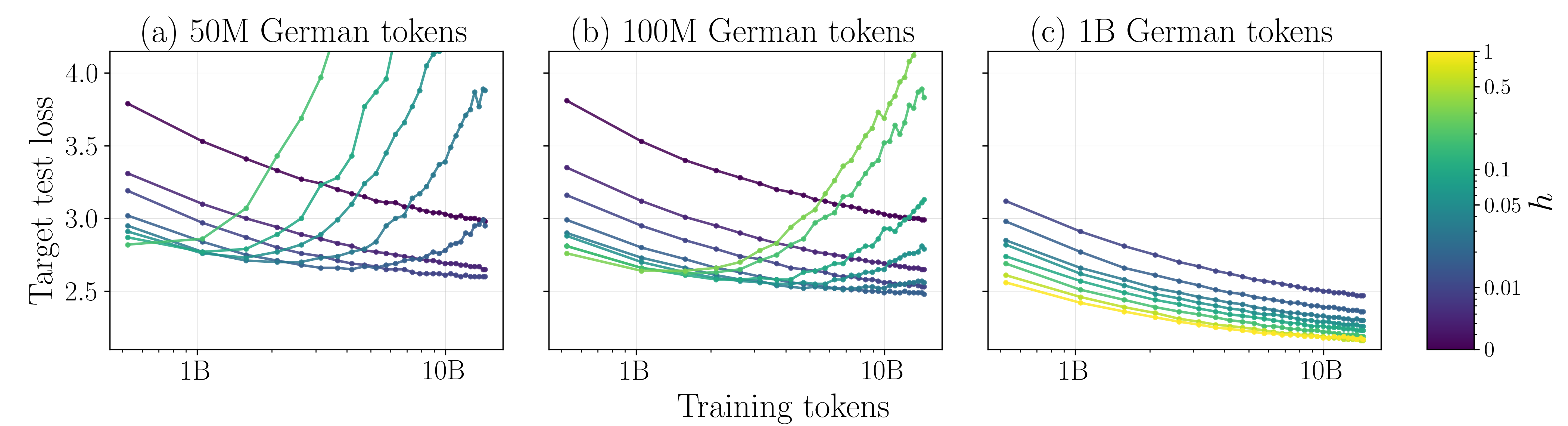}
\vspace{-15pt}
\caption{German validation loss vs.\ total training tokens for the 143M model across three target data budgets. Each curve corresponds to a different target weight $h$.}
\vspace{-10pt}
\label{fig:loss_curves}
\end{figure}
\begin{figure}[!t]
\centering
\includegraphics[width=\textwidth]{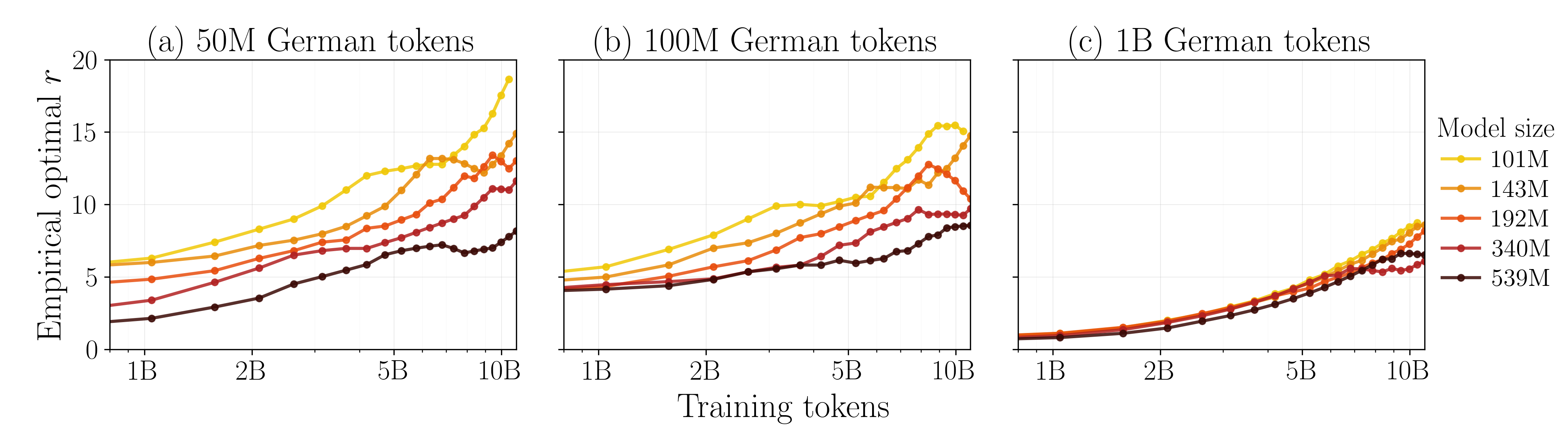}
\vspace{-15pt}
\caption{Empirical optimal $r$ as a function of training tokens for three data constraint sizes across model sizes. Optimal repetition grows steadily with the training budget.}
\label{fig:optimal_r}
\end{figure}
\paragraph{Multi-domain}
The target domain is OpenWebMath \citep{paster2023openwebmath}, a collection of mathematical web pages, constrained to 10M, 100M, or 1B tokens. The generic domain is a subset of DCLM \citep{li2024datacomp}, a large-scale curated English web corpus.
We additionally run three-domain experiments using Wikipedia and peS2o \citep{soldaini2023pes2o}, a corpus of scientific papers derived from Semantic Scholar, both sourced from the OLMo data mix \citep{walsh2024olmo2}, with dataset sizes ranging from 10M-500M and 50M-2.5B, respectively. As the generic domain, we use Fineweb \citep{penedo2024fineweb} and Nemotron-CC \citep{dash2024nemotroncc}.
Evaluation is done on held-out sets from the corresponding datasets. 
\paragraph{Quality}
We score DCLM \citep{li2024datacomp} documents using the original fasttext quality classifier\footnote{\url{https://huggingface.co/mlfoundations/fasttext-oh-eli5}} 
and select the top \{1, 5, 10, 20\}\% of the DCLM base corpus as the \emph{high-quality} (HQ) target set, with the full DCLM base data as the \textit{low-quality} generic set. Unlike the different domain and language experiments, exploring data quality allows us to explore the trade-off of amount of data from the data constrained domain, and closeness to the target domain.  Stricter filtering yields a lower set of HQ data and thus more repetition at a given $h$, but the data is expected to be closer on average to the target set. In contrast, a lower filter percent will yield more data, but will be further from the target set. 
Evaluation is performed on the data used to train the original classifier.

Experiments use GPT-2-style autoregressive decoder-only Transformer language models spanning sizes from 101M to 805M non-embedding parameters (see Appendix~\ref{app:training_details} for full architecture specifications).
Models are trained for approximately $100 \times N$ tokens, where $N$ is the non-embedding parameter count, a sufficient horizon to observe both the benefits and eventual degradation from data repetition.
Data sources are mixed at the sample level: each mini-batch contains examples sampled independently from the target set with probability $h$ and from the generic set with probability $1-h$.

%% file: chapters/04_results.tex
\section{Empirical Findings}
\label{sec:results}

We present results primarily through the German-English bilingual setup due to space constraints,  simulating the low-resource scenario with German.
Results for other languages and domains are consistent and discussed in Appendix~\ref{app:consistency}.

\paragraph{Data Repetition Leads to Predictable Overfitting}

Figure~\ref{fig:loss_curves} shows target-domain loss for the 143M model across three target dataset sizes, with each curve corresponding to a different target weight $h$. When $h$ forces high repetition, loss first improves then rises sharply, reflecting overfitting.
Target datasets of larger size delay this degradation because the same $h$ produces less repetition (Eq.~\ref{eq:main}).
Crucially, across all settings, overfitting onset is governed by the repetition factor $r$ alone: it means, the same target weight can be safe or harmful depending on the available dataset, but the same $r$ consistently triggers degradation regardless of how it is reached.               
This predictability of the same threshold, which holds across data sizes, model scales, and domains, motivates the scaling law in Section~\ref{sec:scaling_law}.

\paragraph{Mixture Training Unlocks High Repetition Tolerance}

Figure~\ref{fig:optimal_r} shows the optimal repetition factor (i.e., the value of $r$ that minimizes target loss) as a function of training tokens. 
Across all settings, optimal $r$ increases steadily with the training budget, reaching up to 15–20 depending on model size and dataset size, substantially exceeding the widely used rule-of-thumb of $<4$ epochs for single-source data-constrained training~\citep{muennighoff2023scaling}.
The difference stems from the generic domain, which provides a continuous supply of fresh tokens that effectively regularizes training and allows the model to absorb far more target-domain repetition before degradation sets in. Since $r$ results from the interplay of $h$, dataset size, and training budget (Eq.~\ref{eq:main}), the target weight needed to reach optimal $r$ varies widely, e.g., from 9.5\% (101M) to 1.9\% (539M) for the 50M dataset.

\begin{wrapfigure}{R}{0.52\textwidth}
\centering
\vspace{-15pt}
\includegraphics[width=0.52\textwidth]{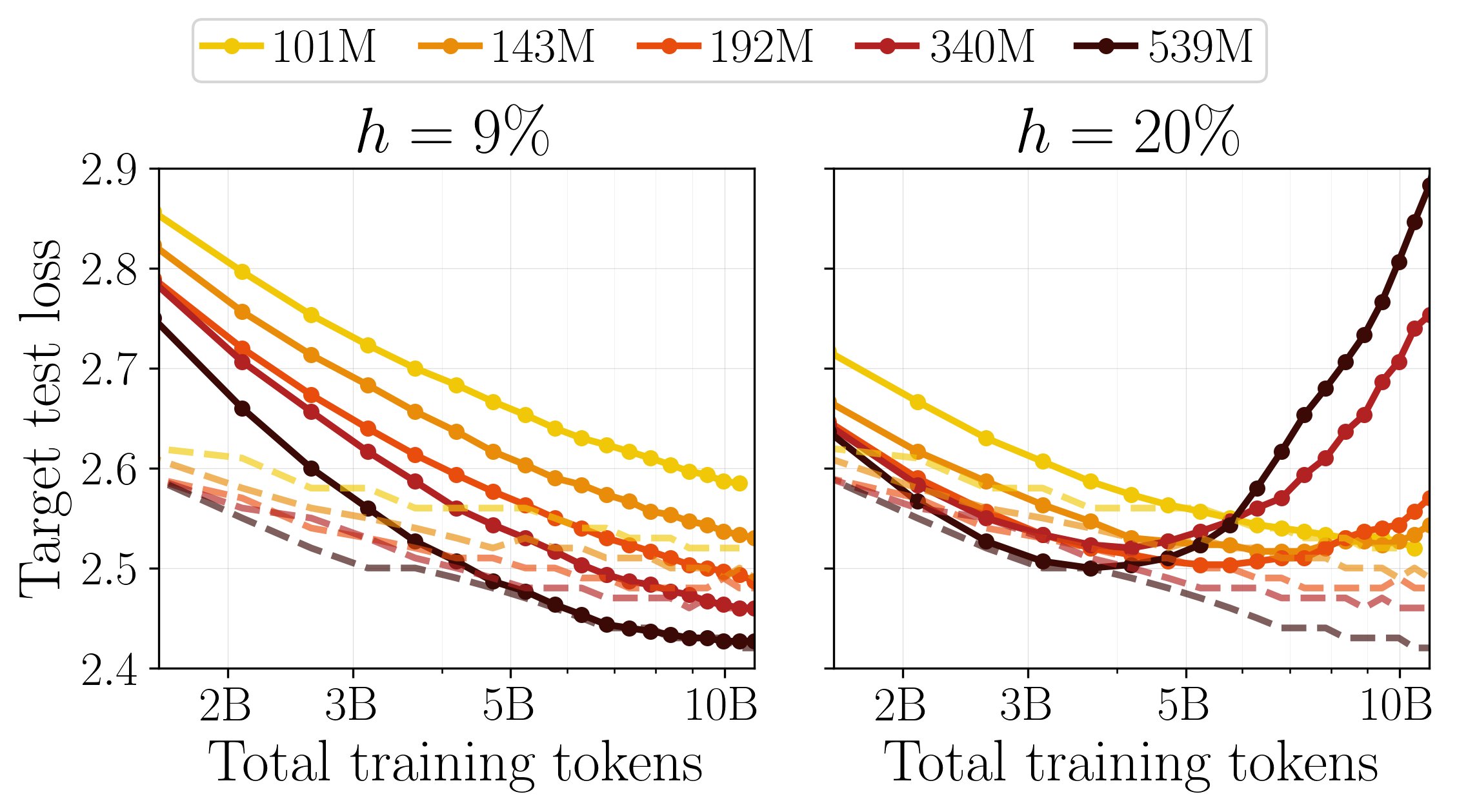}
\caption{Target-domain loss across model sizes at 9\% and 20\% target weights for fixed 100M German budget. Dashed lines show the best-loss envelope across all $h$ values. 
}
\label{fig:cross_model_fixed_h}
\vspace{-10pt}
\end{wrapfigure}

\paragraph{Larger Models Overfit Faster, Yet Still Win}
Figure~\ref{fig:cross_model_fixed_h} shows target-domain loss across five model sizes at two fixed target weights ($h = 9\%$ and $h = 20\%$) with 100M German tokens.
Dashed lines show the best-loss envelope, the minimum achievable loss at each training step across all $h$ values.
At $h = 9\%$, larger models plateau earlier while smaller models are still improving. At $h = 20\%$, the larger models eventually overfit (increasing target loss), while the smallest models plateau or still improve. 
This happens because larger models memorize the data more easily.
Yet in every case, larger models reach lower minima before degradation sets in, and the \emph{best-loss} envelope consistently favors larger models at every training budget.
Thus, we can conclude that \emph{scaling up remains beneficial} even under tight data constraints, but the optimal operating window narrows with model size. This also remains consistent with Figure~\ref{fig:optimal_r}, where the optimal repetition decreases with model size.

\begin{wrapfigure}{R}{0.4\textwidth}
\centering
\includegraphics[width=0.4\textwidth]{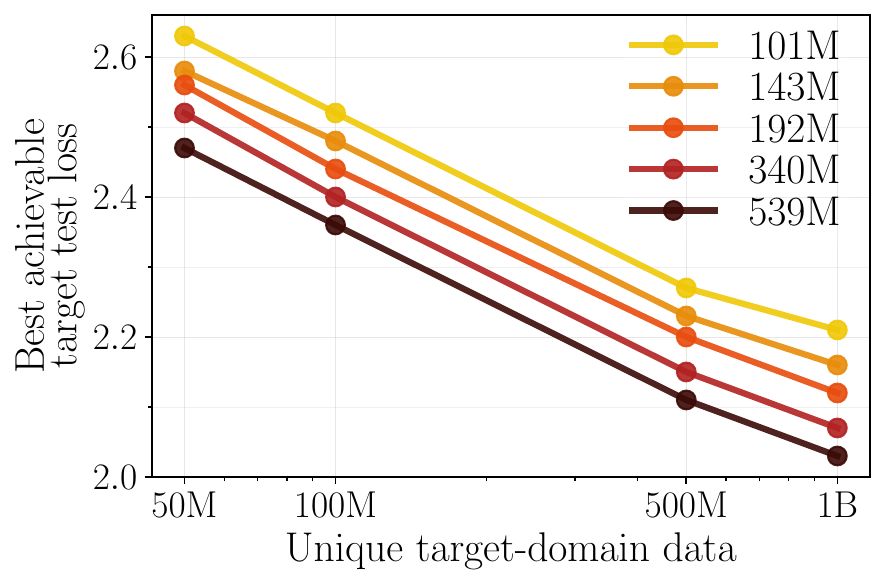}
\vspace{-15pt}
\caption{Best achievable target loss by dataset size across different model sizes with optimal repetition.}
\vspace{-10pt}
\label{fig:diminishing}
\end{wrapfigure}
\paragraph{Predictable Scaling}
Figure~\ref{fig:diminishing} shows \emph{best} achievable loss as a function of unique data available among all weights $h$ and total training tokens, for different model sizes.
We see that these curves are remarkably regular: scaling up model size, or the number of available target data (e.g., by collecting fresh target data) both have a very predictable effect on the loss.

\paragraph{Broad Target Data Filters Can Beat More Repetitions}

The preceding experiments treat the domain dataset as fixed and study how repetition degrades loss. In some settings, however, the dataset size is a design choice as one can choose how to classify documents into the domain of interest. For example, using a quality filter at varying quality levels determines how much data remains, and a practitioner can trade quality for quantity by adjusting the threshold.  We use the Quality Data Mixture to determine whether accepting lower quality data (that is still data constrained and closer to the target domain) to reduce repetition yields better outcomes than repeating the highest quality data\footnote{For many data constrained domains, selecting the amount to filter out is tunable. We study data quality as existing work gives us an already established way of easily varying the amount of data and relevance to the target domain \citet{li2024datacomp}}.

To ground the experiment, we first confirm with a high quality target dataset of the 99th percentile and show loss curves at three dataset sizes (26M, 132M, 264M tokens) (see Figure~\ref{fig:top_1_data_constrained} in Appendix~\ref{app:quality_scales}). At the smallest scale, only $h \leq 0.1$ avoids overfitting, and larger sets progressively unlock higher sustainable target weights. This establishes the baseline that for a fixed quality threshold, the repetition penalty imposes the same constraint on the target loss.

\begin{wrapfigure}{r}{0.35\textwidth}
    \centering
    \vspace{-15pt}
    \includegraphics[width=0.35\textwidth]{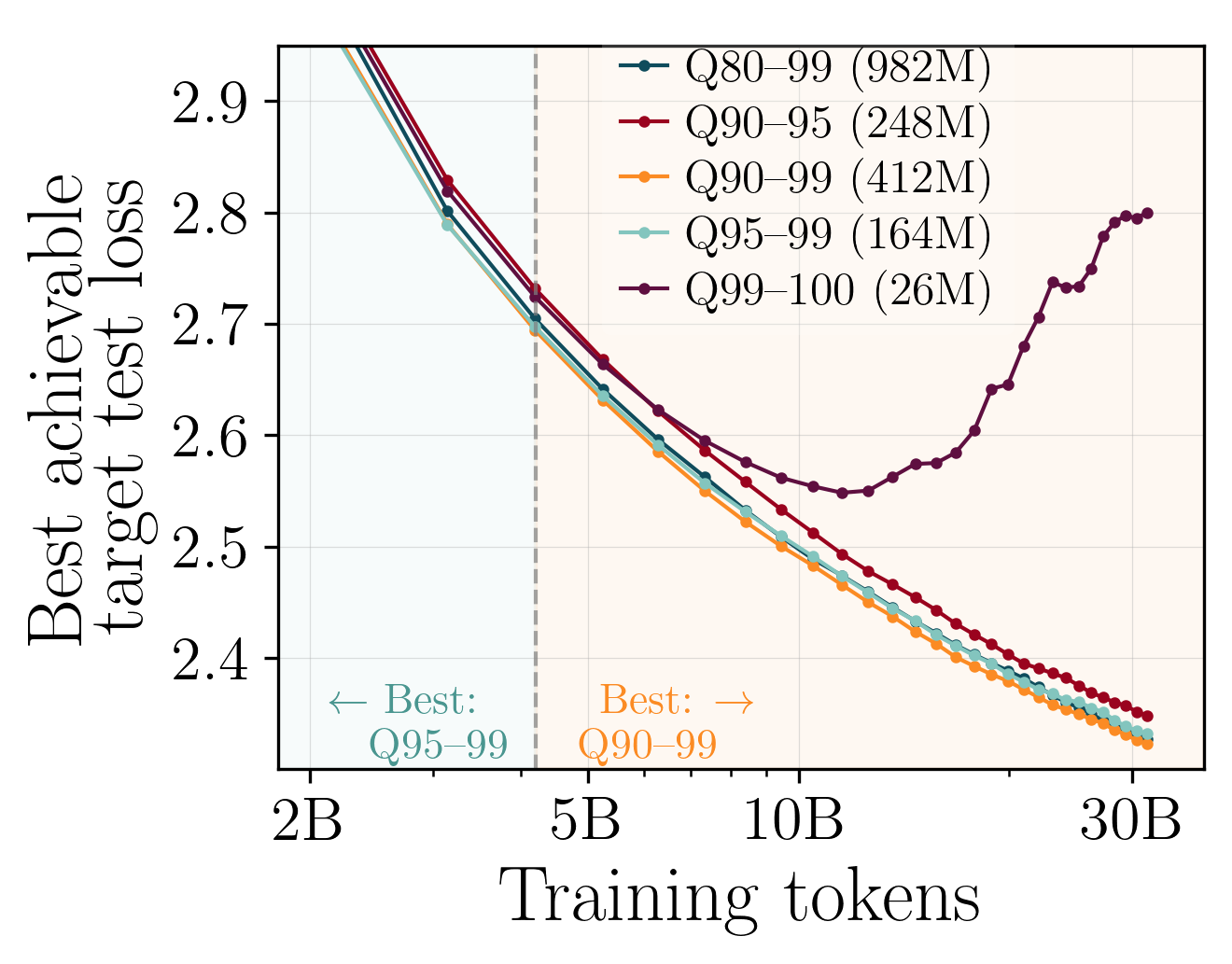}
    \vspace{-20pt}
    \caption{Best loss by quality band. Dashed line marks the crossover where Q90-99 overtakes Q95-99.}
    \label{fig:comparison_best_loss}
    \vspace{-15pt}
\end{wrapfigure}
\begin{figure}[t]
    \centering
    \includegraphics[width=\textwidth]{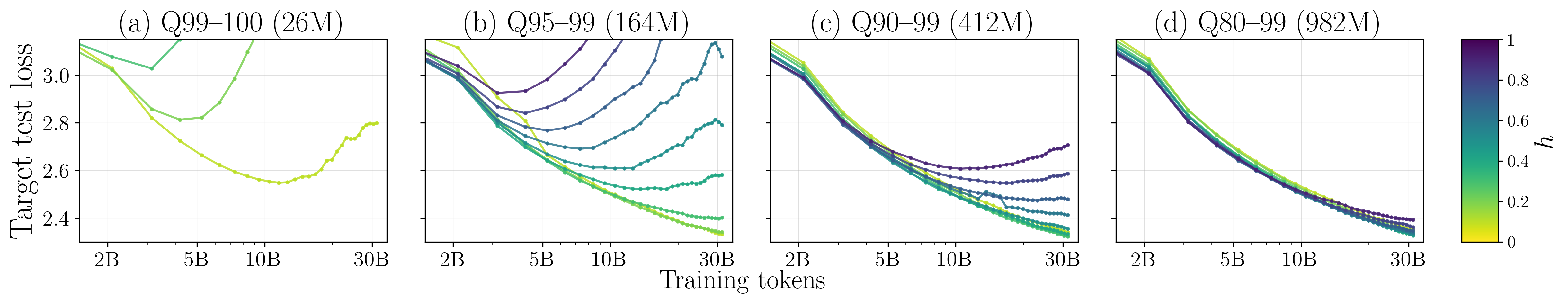}
    \caption{Loss curves for four quality bands ranging from 25M tokens at 99th percentile to 982M at the 80th, and ordered by increasing dataset size. Broader filters enable higher target weights without overfitting.}
    \label{fig:comparison_combined}
    \vspace{-5pt}
\end{figure}

Next, we vary the quality filter threshold while holding the data source fixed. We fix the data constrained set such that the top 1\% (Q99-100) is 26M tokens, and add increasing amounts of data to the data constrained dataset at thresholds of 5\%, 10\%, and 20\% up to 982M tokens.  Results are in Figure~\ref{fig:comparison_combined}.  Relaxing the quality threshold and accepting slightly lower-quality data in exchange for a larger set consistently outperforms repeating a narrow high-quality slice. At Q99-100, only $h = 0.1$ avoids overfitting at up to 10B tokens of training, while Q90-99 supports weights up to $h = 0.4$ and Q80-99 shows nearly all weights monotonically decreasing.

The best quality set also depends on the training budget as we further observe that the best quality filter threshold changes at around 5B tokens as seen in Figure~\ref{fig:comparison_best_loss}, where the top 5\% high quality documents (Q95-99) performs best due to higher per-token quality, but the top 10\% set (Q90-99) overtakes it as training progresses and the smaller dataset saturates.
By 30B tokens, Q90-99 achieves loss 2.33 versus 2.80 for Q99-100 due to the repetition penalty.
This demonstrates that while repetition is an important factor, when high-quality data is scarce, broadening the filter is preferable to over repeating data.

%% file: chapters/05_scaling_law.tex
\section{Repetition-Aware Mixture Scaling Law}
\label{sec:scaling_law}

The goal of this section is to quantify the empirical findings of Section~\ref{sec:results}.
We use scaling laws~\citep{kaplan2020scaling}, which, in their original forms, are simple power laws that allow to predict the train loss of a model from its number of parameters $N$ and total number of tokens it has been trained on $D_\text{total}$.
In our setup, both the target data size $D_\text{target}$ and the target weight $h$ influence the loss on the target domain. 
Our objective is to obtain a simple and predictive formula for $L_\text{target}(N, D_\text{total},  D_\text{target}, h)$.
We focus on the practically relevant regime where the target data is fully consumed at least once, i.e. $r = h \cdot D_\text{total} / D_\text{target} \ge 1$.
When $r < 1$, target tokens are never repeated and mixing reduces to standard mixture selection, as explored, e.g., in \citep{shukor2025scalinglawsoptimaldata}. In the following, we write scaling laws learned parameters in \pp{blue}.

\paragraph{Effective data} 
The core building block of our law is the \emph{effective data} $D_{\mathrm{eff}}$, which accounts for the diminishing value of repeated tokens.
Following \citet{muennighoff2023scaling}, we define the effective contribution of the \emph{target} domain as
\begin{equation}\label{eq:rho}
  D_T = D_\text{target}\bigl(1 + \rho(r)\bigr)\,,
  \qquad \text{with} \quad
  \rho(r) = \pp{r_1}\!\left(1 - e^{-(r-1)/\pp{r_1}}\right),
\end{equation}
where $\pp{r_1}$ is a parameter that controls the effect repeated data.
For small $r \ll \pp{r_1} + 1$, we have $\rho(r) \approx r - 1$ so that $D_T \approx r \cdot D_\text{target}$: each pass counts fully.
For large $r$, $D_T$ saturates at $(1 + \pp{r_1})\,D_\text{target}$, reflecting the diminishing value of further repetitions.
The total effective data is then
\begin{equation}\label{eq:deff}
  D_{\mathrm{eff}} = (1-h)\,D_\text{total} + \pp{\tau}\,D_T\,,
\end{equation}
where the  $(1-h)\,D_\text{total}$ term corresponds to the number of tokens seen from the generic dataset and $\pp{\tau}$ controls the relative value of target-domain tokens compared to generic-domain tokens.
This formulation directly encodes two observations from Section~\ref{sec:results}. The saturation of $\rho(r)$ at $\pp{r_1}$ is the quantitative form of the diminishing value of repeated target tokens: for $r \gg \pp{r_1}$, further passes add essentially nothing to $D_\mathrm{eff}$. Conversely, the unsaturated $(1-h)\,D_\text{total}$ term formalises why mixture training tolerates far more repetition than single-source training (Figure~\ref{fig:optimal_r}): generic tokens never saturate, so they keep $D_\mathrm{eff}$ growing even when the target contribution has plateaued.

\paragraph{Scaling law formulas}
We propose two loss formulas, for fixed and variable model size:
\begin{equation}\label{eq:lsize}
  L_{\mathrm{fix}} = \pp{E} + \frac{\pp{A}}{D_{\mathrm{eff}}^{\,\pp{\alpha}}} + \pp{\gamma}\,h\,,
  \qquad
  L_{\mathrm{size}} = \pp{E} + \frac{\pp{C}}{N^{\pp{\beta}}} + \frac{\pp{B} \, N^{\pp{\delta}}}{D_{\mathrm{eff}}^{\,\pp{\alpha}}} + \pp{\gamma}\,h\,.
\end{equation}
In $L_\mathrm{fix}$, the irreducible loss $\pp{E}$ is a constant baseline; the data term $\pp{A}/D_{\mathrm{eff}}^{\,\pp{\alpha}}$ is a Chinchilla-style power-law relating loss to effective data; and the weight penalty $\pp{\gamma}\,h$ is a linear cost on the target weight.
$L_\mathrm{size}$ additionally includes a Chinchilla-style capacity term $\pp{C}/N^{\pp{\beta}}$ and a data-size coupling $N^{\pp{\delta}}$ ($\pp{\delta} > 0$)~\citep{hoffmann2022training}: the capacity term captures the intrinsic capability gain from scale, while the coupling means that for the same $D_\mathrm{eff}$, the data-limited loss component is amplified at larger~$N$, encoding the fact that larger models yield better losses but overfit faster (Figure~\ref{fig:cross_model_fixed_h}).
At fixed $N$, $L_\mathrm{size}$ reduces exactly to $L_\mathrm{fix}$ with $\pp{A} = \pp{B}\,N^{\pp{\delta}}$ and $E_{\mathrm{fix}} = \pp{E} + \pp{C}/N^{\pp{\beta}}$, ensuring consistency.
$L_\mathrm{fix}$ has six fitted parameters ($\pp{E}$, $\pp{A}$, $\pp{\alpha}$, $\pp{r_1}$, $\pp{\tau}$, $\pp{\gamma}$); $L_\mathrm{size}$ has nine ($\pp{E}$, $\pp{C}$, $\pp{\beta}$, $\pp{B}$, $\pp{\delta}$, $\pp{\alpha}$, $\pp{r_1}$, $\pp{\tau}$, $\pp{\gamma}$).

\paragraph{Fitting}
Given a collection of training runs indexed by $i$, each characterised by a total token budget $D_{\text{total},i}$, a target weight $h_i$, an available target dataset $D_{\text{target},i}$, and optionally a model size $N_i$, we observe the resulting loss on the target domain $\ell_i$.
The parameters $\theta$ of the scaling law are estimated by minimising the reweighted Huber loss
\begin{equation}\label{eq:fit}
  \hat\theta = \arg\min_\theta \sum_i \omega_i \cdot \mathcal{H}\!\bigl(\ell_i - L_\theta(D_{\text{total},i},\, h_i,\, D_{\text{target},i},\, N_i)\bigr)\,,
\end{equation}
where $\mathcal{H}$ is the Huber loss and the weights $\omega_i = \max(r_i \cdot h_i,\, \epsilon)$ with $\epsilon=0.01$ emphasise the high-repetition, high-fraction regime: $h$ alone would under-weight high-repetition runs at moderate fractions, while $r$ alone would under-weight high-fraction runs with large datasets (where $r$ is low).
The qualitative ranking of methods is unchanged under alternative weights; see Appendix~\ref{app:weighting}.
Following~\citet{shukor2025scalinglawsoptimaldata}, we minimise~\ref{eq:fit} using basin-hopping optimisation with 100 random restarts to avoid poor local minima in the non-convex loss landscape.

\section{Scaling Law Results}
\label{sec:scaling_results}
We evaluate the scaling law on experimental setups described in Section~\ref{sec:datasets}.
For each setup, $L_\mathrm{fix}$ is fitted independently per model size, and $L_\mathrm{size}$ is fitted jointly across all model sizes.
To test extrapolation for $L_\mathrm{fix}$, we fit on the first 50\% of training steps and test on second half. For $L_\mathrm{size}$, we fit on all but the largest model scale, and test on the held-out largest model scale.

\begin{table}[t]
\centering
\caption{Test weighted $R^2$. \textbf{Left:} fixed-size formulas, fitted independently per model size. \textbf{Right:} multi-size formulas, fitted on smaller model sizes and evaluated on the held-out largest model.}
\label{tab:scaling_r2}\label{tab:scaling_r2_size}
\setlength{\tabcolsep}{3pt}
\begin{tabular}{lcccc|lcc}
\toprule
& German & Maths & Quality & Wiki/peS2o & & German & Maths \\
\midrule
$L_\mathrm{fix}$ & \textbf{0.95} & \textbf{0.88} & \textbf{0.71} & \textbf{0.80} & $L_\mathrm{size}$ & \textbf{0.65} & \textbf{0.73} \\
Repetition-agnostic & 0.78 & 0.78 & 0.14 & 0.72 & Repetition-agnostic+$N$ & 0.59 & 0.71 \\
Utility decay & 0.72 & 0.55 & $-$0.64 & 0.79 & Utility decay+$N$ & 0.56 & 0.69 \\
Domain-agnostic & $-$40.7 & $-$0.49 & $-$2.19 & $-$1.16 & Domain-agnostic+$N$ & $-$0.23 & $-$0.77 \\
\bottomrule
\vspace{-15pt}
\end{tabular}
\end{table}

\paragraph{Baselines}
We compare against three other scaling laws formulas inspired by the literature.
\emph{Repetitions-agnostic}~\citep{shukor2025scalinglawsoptimaldata} replaces $D_\mathrm{eff}$ with $(1-h)\,D_\text{total} + \pp{\tau}\,h\,D_\text{total}$, treating repeated tokens as unique.
\emph{Domain-agnostic}~\citep{muennighoff2023scaling} uses a single saturating function of total tokens without distinguishing domains.
\emph{Utility decay}~\citep{goyal2024scaling} models repetition through a decaying exponent on the data term.
The formulas are detailed in  Appendix~\ref{app:details}.

\paragraph{Optimal mixture prediction}
\begin{wrapfigure}{r}{0.38\textwidth}
\centering
\includegraphics[width=0.37\textwidth]{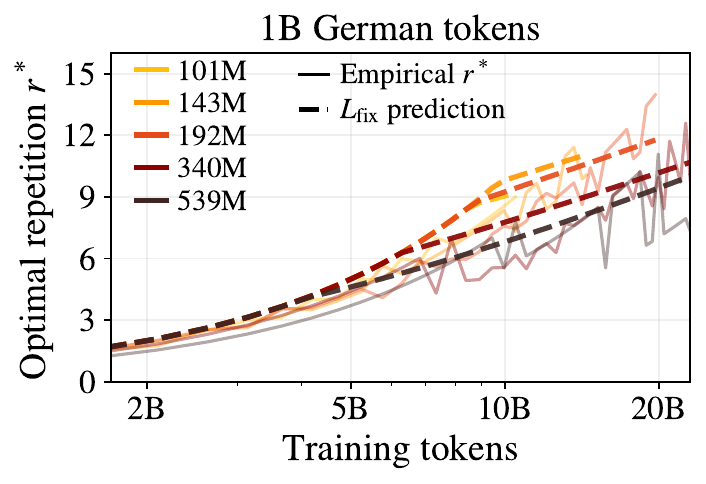}
\caption{Predicted vs.\ empirical optimal repetition $r^*$ for the  German target dataset of 1B tokens. Solid: empirical optimum; dashed: $L_\mathrm{fix}$ prediction.}
\label{fig:predicted_vs_actual_r}
\vspace{-25pt}
\end{wrapfigure}

\paragraph{Loss prediction}
Table~\ref{tab:scaling_r2} reports the weighted coefficient of determination ($wR^2$) on the held-out test split (train $wR^2$ in Appendix~\ref{app:extended_scaling}), where each observation is weighted by $\max(r \cdot h,\, \epsilon)$ to emphasise the high-repetition regime.
$L_\mathrm{fix}$ consistently achieves the best test $wR^2$, extrapolating beyond the fitting range.
\emph{Repetitions-agnostic} fits the training range reasonably but fails to extrapolate to high repetitions. \emph{Domain agnostic} fails completely: treating all tokens as interchangeable does not capture the loss dynamics.
\emph{Utility decay} performs honorably on several datasets.

\begin{wrapfigure}{r}{0.35\textwidth}
\vspace{-10pt}
\centering
\footnotesize
\setlength{\tabcolsep}{4pt}
\begin{tabular}{lccc}
\toprule
& Median & Mean & p90 \\
\midrule
$L_\mathrm{fix}$ & \textbf{26} & \textbf{34} & \textbf{76} \\
Rep-agn. & 88 & 73 & 99 \\
Util. dec. & 31 & 41 & 95 \\
Dom.-agn. & 47 & 59 & 98 \\
\bottomrule
\end{tabular}
\captionof{table}{Fraction of training tokens wasted by following each formula's predicted optimal mixture instead of the oracle.}
\label{tab:scaling_weight}
\vspace{-10pt}
\end{wrapfigure}
The primary use of the scaling law is to predict the optimal target weight~$h^*$ for a given total token budget $D_\text{total}$ and available target domain dataset $D_\text{target}$.
Once the scaling law parameters are estimated, we can simply solve
\begin{equation}\label{eq:hstar}
  h^* = \arg\min_{h \in [0,1]} L_\mathrm{fix}(D_\text{total},\, h,\, D_\text{target})
\end{equation}
by grid search over~$h$. Figure~\ref{fig:predicted_vs_actual_r} shows an example of empirical vs. estimated optimal repetition.

To quantify the practical cost of suboptimal mixture predictions, we measure the fraction of training tokens \emph{wasted} by following each formula's recommendation instead of the oracle (Table~\ref{tab:scaling_weight}).
At a given budget $D_\text{total}$, the formula recommends a fraction~$h^*_\text{pred}$; we interpolate the empirical loss at that fraction and find the budget $D'$ at which the run using the empirically best fraction at each step first reached the same loss.
The wasted fraction $(D_\text{total} - D') / D_\text{total}$ directly measures how much compute could have been saved with perfect knowledge of the optimal mixture.
Detailed results are in Appendix~\ref{app:extended_scaling}.

%% file: chapters/06_ablations.tex
\section{Multiple Domain Experiments}
\label{sec:multi_domain}

\begin{figure*}[t]
   \centering
   \includegraphics[width=\textwidth]{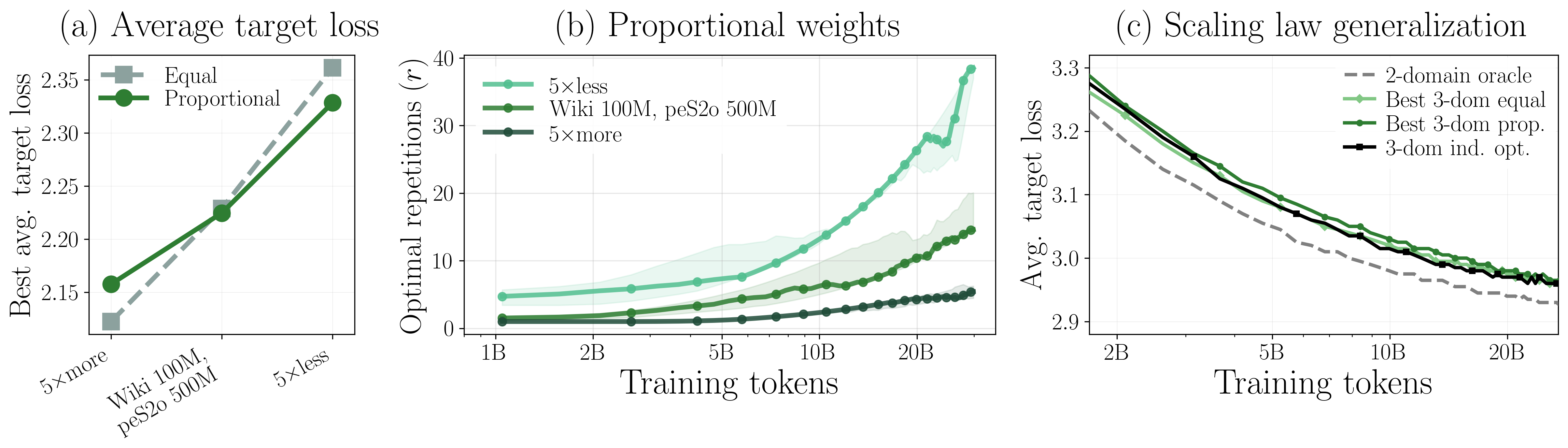}
   \vspace{-15pt}
   \caption{Multi-domain experiments (101M model, English + Wikipedia + peS2o). \textbf{(a)} Best achievable average target loss under equal vs.\ proportional weighting across three set-size configurations. \textbf{(b)} Optimal repetitions under proportional weighting with 10\% compute confidence band (shaded). \textbf{(c)} Independently optimizing $r$ per domain using the bilingual scaling law outperforms the best grid-searched proportional weighting.}
   \vspace{-10pt}
   \label{fig:3domain_combined}
\end{figure*}

In order to test whether our findings and scaling law formula extend to settings with more than one constrained domain, we set up experiments with a mixture of three domains: one unconstrained generic source and two constrained target domains.
We use two configurations: FineWeb as the generic source with peS2o and Wikipedia as targets, and Nemotron as the generic source with Wikipedia and peS2o as targets.\footnote{Subplot~(a) of Figure~\ref{fig:3domain_combined} uses FineWeb + peS2o + Wikipedia; subplots~(b) and~(c) use Nemotron + Wikipedia + peS2o.}
We vary the dataset sizes across three configurations: 5$\times$ less, base, and 5$\times$ more data.
Our primary metric is the average loss across both target domains, analogous to evaluating on a test set that combines text from both domains.

\paragraph{Weighting Strategies}
First, we compared two weighting strategies: \emph{equal weighting} ($h_{\text{wiki}} = h_{\text{peS2o}}$) and \emph{proportional weighting}, where fractions are set proportional to dataset size ($h_{\text{wiki}} / h_{\text{peS2o}} = D_{\text{wiki}} / D_{\text{peS2o}}$).
Our experiments demonstrate (Figure~\ref{fig:3domain_combined}a) that by average loss, proportional weighting outperforms equal weighting when data is scarce, while equal weighting is better in the data-rich regime. 
The two strategies cross near the base configuration, suggesting that proportional weighting becomes increasingly beneficial as repetition grows. 
For the rest of this section, we focus on the proportional weighting, and provide the results for equal weighting in Appendix~\ref{app:3domain}. 

\paragraph{Empirical Findings}
The key findings from the bilingual experiments carry over to the multi-domain setting.
Figure~\ref{fig:3domain_combined}b shows that optimal repetition grows steadily with training tokens across all set-size configurations, with smaller sets requiring substantially higher repetition: the 5$\times$less variant reaches $r \approx 30$ while 5$\times$more stays below $r \approx 7$. This mirrors the bilingual experiments: the generic domain continues to regularize training even with multiple constrained domains present, enabling high repetition tolerance. 
Importantly, the choice of $r$ need not be exact. We quantify this through a 10\% compute confidence band (shaded regions in Figure~\ref{fig:3domain_combined}b): at each checkpoint, we identify all values of $r$ that achieve loss no worse than the best loss achievable with 10\% less compute. The bands are wide, particularly for smaller sets, indicating that even rough estimates of optimal $r$ yield near-optimal target loss.

\paragraph{Scaling Law Generalization}
The similarity of these findings to the bilingual setting raises the question of whether the scaling law transfers directly to multi-domain mixtures. We test this by setting each domain's repetition factor independently using the optimal $r$ predicted by the bilingual scaling law for the corresponding set size ($r_\text{wiki}=35$, $r_\text{peS2o}=17$). Figure~\ref{fig:3domain_combined}c shows that this independently optimized configuration consistently outperforms the best proportional weighting and narrows the gap to the 2-domain oracle (the average loss achieved by training each domain separately with English). While a gap to the oracle remains, this result suggests that costly multi-domain sweeps over all possible combinations of domain weights can be avoided: our two-domain scaling law can independently estimate the optimal $r$ for each constrained domain, and these can be combined into a single mixture, replacing an entire experimental grid with a single, better-performing run.

%% file: chapters/07_conclusion.tex
\section{Conclusion}
\label{sec:conclusion}

We studied mixture pretraining under data constraints across diverse data types: multilingual, multi-domain, and quality-filtered.
Our experiments reveal consistent patterns across all settings: repeated target-domain tokens yield diminishing returns, optimal repetition scales predictably with data availability, and larger models extract more from limited data.
Crucially, mixture training tolerates substantially higher repetition than single-source training, with generic data acting as an implicit regularizer.
We formalized these dynamics in a scaling law that accurately predicts the optimal mixture across experimental setups, outperforming baselines that ignore either repetition or domain structure.
In practice, this allows training on scarce target data without expensive sweeps: given only the target pool size and compute budget, the law prescribes the mixture ratio that maximizes target-domain performance in a single run.

%% file: chapters/09_appendix.tex
\section{Experimental Setup}
\label{app:exp_details}

\paragraph{Multilingual setup}
For each combination of model size and pool size, we sweep $h$ over a fine grid of 19--27 values (see Table~\ref{tab:exp_setup} for a summary), with denser sampling at low repetition levels where the loss landscape changes most rapidly. The grid spans number of repetitions from below 1 to above 30, though configurations where the resulting $h$ exceeds 1 are excluded.
For larger models, the feasible range of $h$ is narrower because the longer training duration ($100 \times N$ tokens) produces higher repetition at the same $h$; for example, the 539M model on the 50M pool covers $r$ up to approximately 20.
Evaluation is performed on 10,000 samples from held-out test sets of both FineWeb2 (target language) and FineWeb (English), allowing us to measure both target-domain and generic-domain loss at each checkpoint.

\paragraph{Multi-domain setup}
For the three-domain experiments (Section~\ref{sec:multi_domain}), we train 101M models with a total training budget of 30B tokens (approximately $300 \times N$), increased relative to the bilingual setup to allow sufficient repetition of both target domains.
The three pool-size configurations correspond to: 5$\times$less (Wiki 20M, peS2o 100M), base (Wiki 100M, peS2o 500M), and 5$\times$more (Wiki 500M, peS2o 2.5B).
We sweep over total target fractions $h_{\text{total}} = h_{\text{wiki}} + h_{\text{peS2o}}$ ranging from 0.02 to 0.60 with step 0.02, under both equal weighting ($h_{\text{wiki}} = h_{\text{peS2o}}$) and proportional weighting ($h_{\text{wiki}} / h_{\text{peS2o}} = D_{\text{wiki}} / D_{\text{peS2o}}$).

\paragraph{Math setup}
For the OpenWebMath experiments, we sweep $h$ over a linear grid whose density depends on pool size: 26 values for the 10M and 100M pools and 8 values for the 1B pool (60 total per model size). The 805M model uses a denser grid (35, 35, and 20 values for the 10M, 100M, and 1B pools respectively; 90 total), because its longer training horizon (1M vs.\ 200K steps) produces higher repetition counts that require finer resolution.
The $h$ range is adapted to each pool size, spanning up to $h = 0.078$ for the 10M pool, $h = 0.78$ for the 100M pool, and $h = 1.0$ for the 1B pool.
The maximum number of repetitions reached is approximately 41 for the 130--630M models and 92 for 936M.
The 101M--539M models are trained for 200K steps (13.1B tokens) and the 805M model for 1M steps (65.5B tokens), with a reduced batch size of 32 sequences. The resulting tokens-to-parameters ratios range from approximately $130\times$ (101M) down to $24\times$ (539M), with $81\times$ for 805M.
These experiments use a SentencePiece tokenizer \citep{kudo2018sentencepiece} trained on C4 \citep{raffel2020exploring} with a vocabulary of 32,000 tokens, rather than the GPT-2 BPE tokenizer used in other setups.
Evaluation is performed on 10,240 sequences (10.5M tokens) held out from OpenWebMath.

\paragraph{Quality setup}
For the quality filter experiments, we train 100M parameter models ranging from 1--30B tokens, sweeping over $h \in \{0.1, 0.2, \dots, 1.0\}$.  We train models with varying quality pools primarily either 1\%, 5\%, and 10\% with non-overlapping bins between the 1\% and 5\% and 10\% runs to explicitly measure the impact of having lower quality data, but with lower repetition.

\section{Training Details}
\label{app:training_details}

\paragraph{Model architecture}
All models are GPT-2-style autoregressive decoder-only Transformers with learned positional embeddings.
Models are scaled along a depth-based rule where the hidden dimension $d_\text{model} = 128 \times L$ for $L$ layers, with an FFN intermediate size of $4 \times d_\text{model}$, attention head dimension of 64, and $d_\text{model}/64$ attention heads.
All models use a context length of 1,024 tokens.
Further details are provided in Table~\ref{tab:model_arch}.

\paragraph{Training}
Multilingual models are trained for approximately $100 \times N$ tokens, where $N$ is the non-embedding parameter count.
The three-domain experiments are trained for $300 \times N$, the quality experiments are trained for 30K steps; the details about the OpenWebMath experiments are provided above.
All models are optimized with Adam \citep{kingma2014adam}, 
a constant learning rate following a linear warmup, weight decay of $10^{-2}$, gradient clipping at 0.1, and no dropout.
The multilingual and Wiki/peS2o experiments use a learning rate of $10^{-3}$ (selected based on initial experiments for 101M, 143M, and 192M models on English-German data), a warmup of 1\% of training steps, a batch size of 256 sequences, and a GPT-2 BPE tokenizer \citep{radford2019language} with a vocabulary of 50,257 tokens.
The quality experiments use a learning rate of $3 \times 10^{-4}$, a warmup of 1\% of training steps, a batch size of 1,024 sequences, and the same GPT-2 BPE tokenizer.
The OpenWebMath experiments use a learning rate of $10^{-4}$ with $\mu$P,%
a 2,000-step warmup, a batch size of 32 sequences, and a SentencePiece tokenizer \citep{kudo2018sentencepiece} trained on C4 \citep{raffel2020exploring} with a vocabulary of 32,000 tokens.

\paragraph{Learning Rate}
We use a constant learning rate (after warmup) rather than a cosine or linear decay schedule.
Constant or near-constant schedules are common practice in pretraining work \citep{hagele2024scaling,porian2025resolving}.
Our experimental design also requires it. The central object of analysis is a loss-vs-repetition curve read along a single training run (e.g., Figures~\ref{fig:loss_curves},~\ref{fig:cross_model_fixed_h}). A decay schedule couples the learning rate to the training horizon, so checkpoints at different repetition counts would not be directly comparable within a run. Replicating our experiments under a decay schedule would require a separate run for each evaluation point, multiplying compute by the number of checkpoints.

\paragraph{Compute resources}
All experiments were run on a mix of A100 and H100 GPUs.
A single training run requires approximately 40 GPU-hours for the 101M model, 64 for 143M, 100 for 192M, 264 for 340M, 600 for 539M, and 1200 GPU-hours for 805M.

\begin{table}[t]
\centering
\caption{Model architectures. Parameters refer to non-embedding parameters. 
}
\label{tab:model_sizes}
\begin{tabular}{lrrrr}
\toprule
Params & Layers & Hidden & FFN & Heads \\
\midrule
101M & 8  & 1024 & 4096 & 16 \\
143M & 9  & 1152 & 4608 & 18 \\
192M & 10 & 1280 & 5120 & 20 \\
340M & 12 & 1536 & 6144 & 24 \\
539M & 14 & 1792 & 7168 & 28 \\
805M & 16 & 2048 & 8192 & 32 \\
\bottomrule
\end{tabular}
\label{tab:model_arch}
\end{table}

\begin{table}[h!]
\centering
\caption{Experimental setup summary. $D_\text{target}$ denotes the range of target data pool sizes; \#\,$h$ is the total number of target fraction configurations per model size.}
\label{tab:exp_setup}
\begin{minipage}[t]{0.48\textwidth}
\centering
\begin{tabular}{clcr}
\toprule
& $N$ & $D_\text{target}$ & \# $h$ \\
\midrule
\multirow{17}{*}{\rotatebox[origin=c]{90}{\textit{Multilingual}}}
& \multicolumn{3}{l}{\textit{German}} \\
& 101M & 50M--1B & 72 \\
& 143M & 50M--1B & 75 \\
& 192M & 50M--1B & 75 \\
& 340M & 50M--1B & 77 \\
& 539M & 50M--1B & 82 \\
\cmidrule{2-4}
& \multicolumn{3}{l}{\textit{French}} \\
& 101M & 50M--500M & 63 \\
& 143M & 50M--500M & 64 \\
& 192M & 50M--500M & 63 \\
& 340M & 50M--500M & 55 \\
& 539M & 50M--500M & 57 \\
\cmidrule{2-4}
& \multicolumn{3}{l}{\textit{Swahili}} \\
& 101M & 50M--500M & 63 \\
& 143M & 50M--500M & 64 \\
& 192M & 50M--500M & 63 \\
& 340M & 50M--500M & 55 \\
& 539M & 50M--500M & 57 \\
\bottomrule
\end{tabular}
\end{minipage}
\hfill
\begin{minipage}[t]{0.48\textwidth}
\centering
\begin{tabular}{clcr}
\toprule
& $N$ & $D_\text{target}$ & \# $h$ \\
\midrule
\multirow{8}{*}{\rotatebox[origin=c]{90}{\textit{Domain}}}
& \multicolumn{3}{l}{\textit{Math (OpenWebMath)}} \\
& 101M & 10M--1B & 60 \\
& 197M & 10M--1B & 60 \\
& 340M & 10M--1B & 60 \\
& 539M & 10M--1B & 60 \\
& 805M & 10M--1B & 90 \\
\cmidrule{2-4}
& \multicolumn{3}{l}{\textit{Wiki + peS2o (3-domain)}} \\
& 101M & 10M--500M & 153 \\
\midrule
\multirow{2}{*}{\rotatebox[origin=c]{90}{\textit{Qual.}}}
& \multicolumn{3}{l}{\textit{Quality filtering}} \\
& 101M & 25M--264M & 10 \\
\bottomrule
\end{tabular}
\end{minipage}
\end{table}

\newpage
\section{Consistency of the Findings Across Languages and Domains}
\label{app:consistency}
The core findings from the German--English bilingual experiments generalize across all settings tested.
We replicate the experimental grid for French and Swahili, which differ substantially from German in morphology, script frequency, and available web data.
We observe qualitatively identical behavior in all cases. 
The same overfitting emerges at high $h$ with small data pools, the same dependence of optimal $r$ on pool size holds, and larger models consistently achieve lower best-case loss despite earlier overfitting onset. We further verify these patterns on non-multilingual but domain setup (Wikipedia, peS2o, OpenWebMath mixed with generic English data), confirming that the repetition dynamics are a general property of mixture training under data constraints rather than an artifact of any particular domain or language pair.

Below we provide results of our experiments organized as follows:
\begin{itemize}
    \item \textbf{Optimal repetition} (Section~\ref{app:optimal_r}): French, Swahili, and OpenWebMath all exhibit the same steady growth of optimal $r$ with training budget, governed by data pool size rather than specific domain.
    \item \textbf{Loss curves} (Section~\ref{app:loss_curves}): Full loss-vs-tokens curves for all model sizes and languages confirm the overfitting at high $h$ with small pools and monotonic improvement with large pools.
    \item \textbf{Cross-model behavior} (Section~\ref{app:cross_model}): Larger models overfit faster yet achieve lower minima across all languages (confirming the pattern we observed for German).
    \item \textbf{Quality experiments} (Section~\ref{app:quality_scales}): The repetition--diversity trade-off extends to quality-filtered data at multiple data scales.
    \item \textbf{Comparison to unlimited data} (Section~\ref{app:unlimited}): When target data is unconstrained, mixture training provides no benefit over single-domain training, confirming that our findings are specific to the data-constrained regime.
\end{itemize}

\section{Optimal Repetition Across Languages and Domains}
\label{app:optimal_r}

\begin{figure}[h!]
\centering
\vspace{-10pt}
\includegraphics[width=\textwidth]{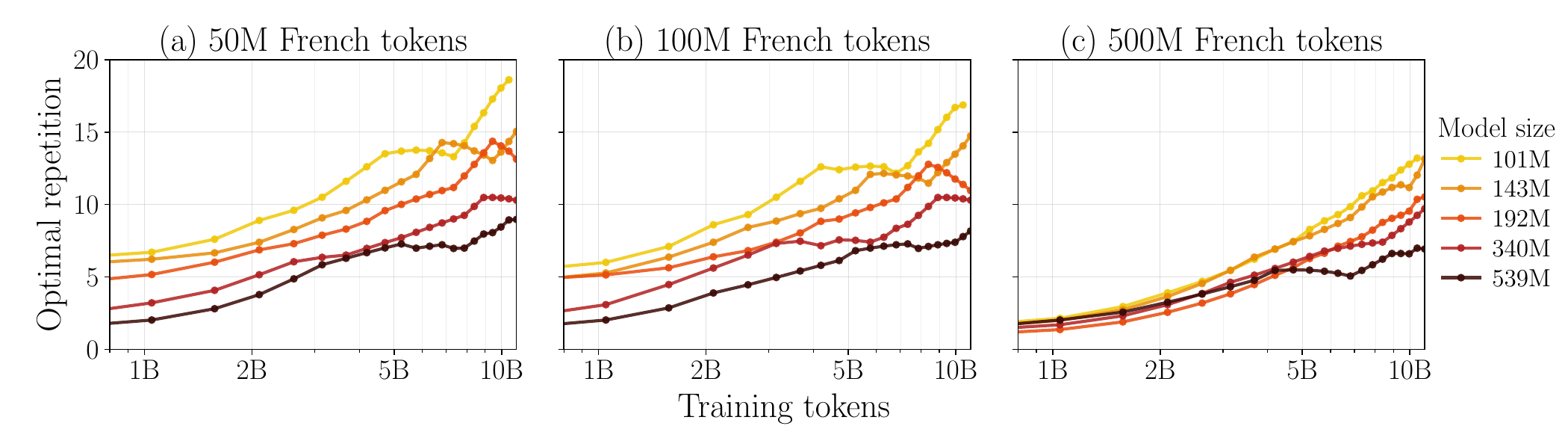}
\caption{Optimal repetition as a function of training tokens for three data constraint sizes across different model sizes (French).}
\label{fig:app_optimal_r_fr}
\end{figure}
\begin{figure}[h!]
\centering
\vspace{-10pt}
\includegraphics[width=\textwidth]{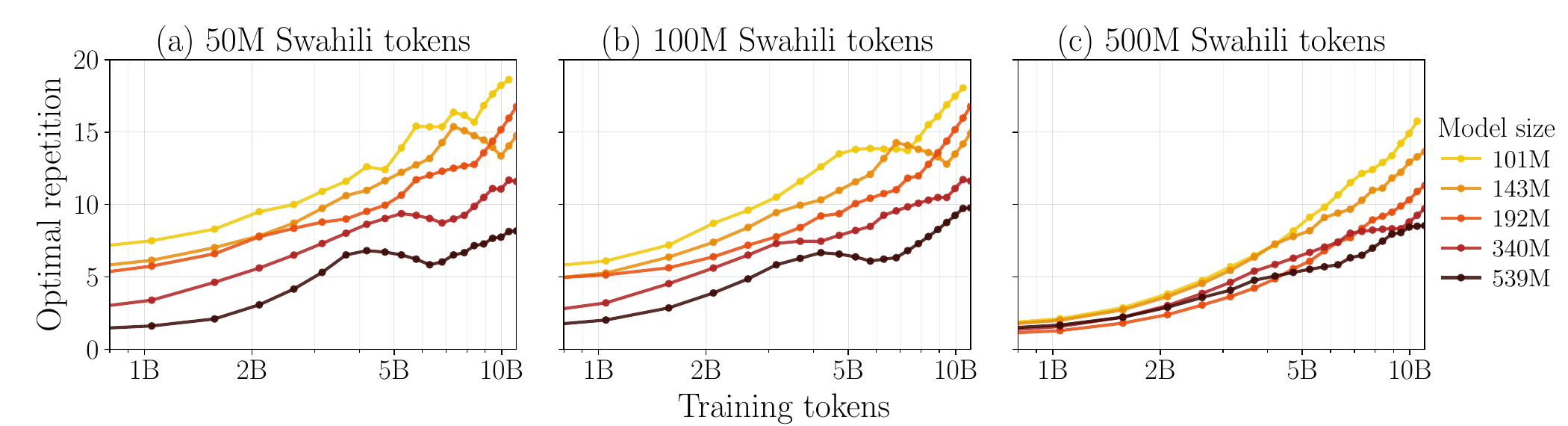}
\caption{Optimal repetition as a function of training tokens for three data constraint sizes across different model sizes (Swahili).}
\label{fig:app_optimal_r_sw}
\end{figure}

Figures~\ref{fig:app_optimal_r_fr}--\ref{fig:app_optimal_r_sw} show the optimal number of repetitions as a function of training tokens for French and Swahili, matching the German plot in the main text (Figure~\ref{fig:optimal_r}). The same steady growth of optimal $r$ with training budget is observed across all languages.

\begin{figure}[h!]
\centering
\includegraphics[width=\textwidth]{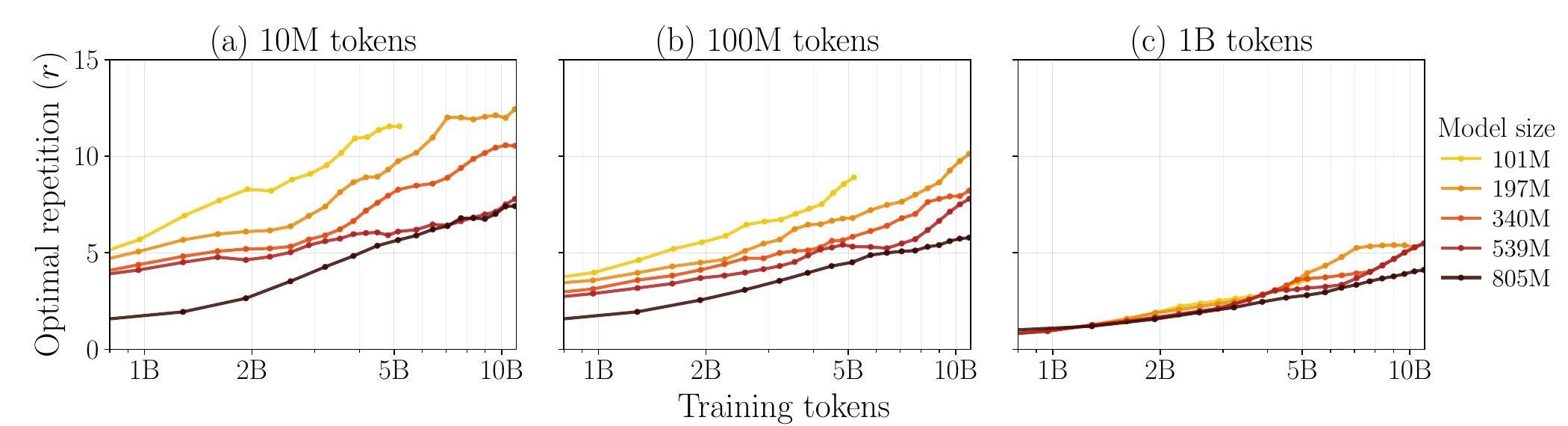}
\caption{Optimal repetition as a function of training tokens for three data constraint sizes across different model sizes (OpenWebMath).}
\label{fig:app_optimal_r_math}
\end{figure}

Figure~\ref{fig:app_optimal_r_math} further extends this analysis to a non-language domain: OpenWebMath mixed with DCLM as the generic source. Despite the very different nature of mathematical text, the same qualitative pattern holds: optimal $r$ increases with training tokens, decreases with pool size, and smaller models tolerate higher repetition (since they are trained on more tokens relative to their capacity). This confirms that the repetition dynamics are a general property of mixture training under data constraints, independent of the specific domain or language.

\newpage
\section{Loss Curves Across Languages, Model Sizes, and Domains}
\label{app:loss_curves}

\begin{figure}[h!]
\centering
\includegraphics[width=\textwidth]{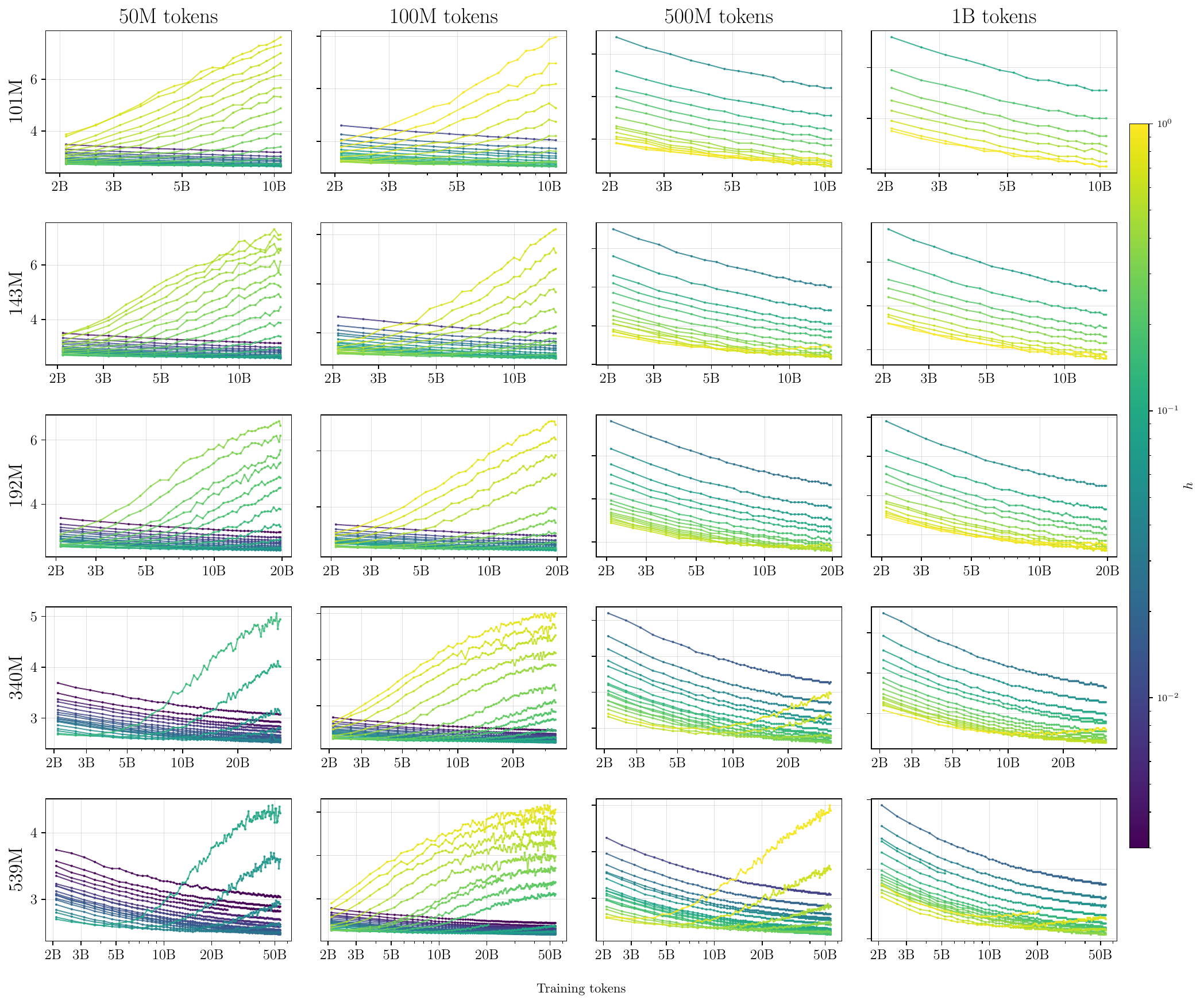}
\caption{German target-domain loss curves across all model sizes (rows) and data pool sizes (columns).}
\label{fig:app_loss_de}
\end{figure}

\begin{figure}[h!]
\centering
\includegraphics[width=\textwidth]{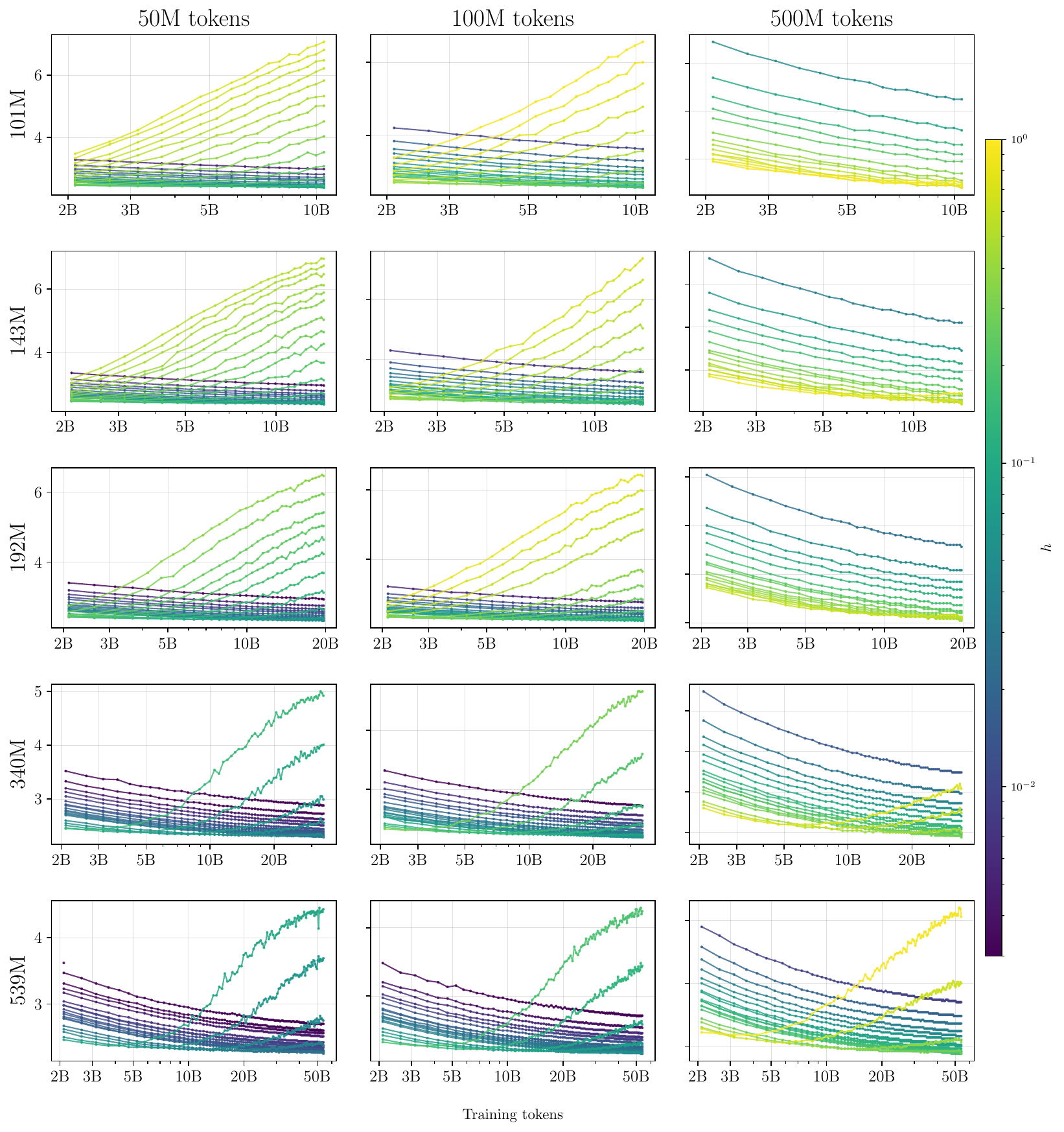}
\caption{French target-domain loss curves across all model sizes (rows) and data pool sizes (columns).}
\label{fig:app_loss_fr}
\end{figure}

\begin{figure}[h!]
\centering
\includegraphics[width=\textwidth]{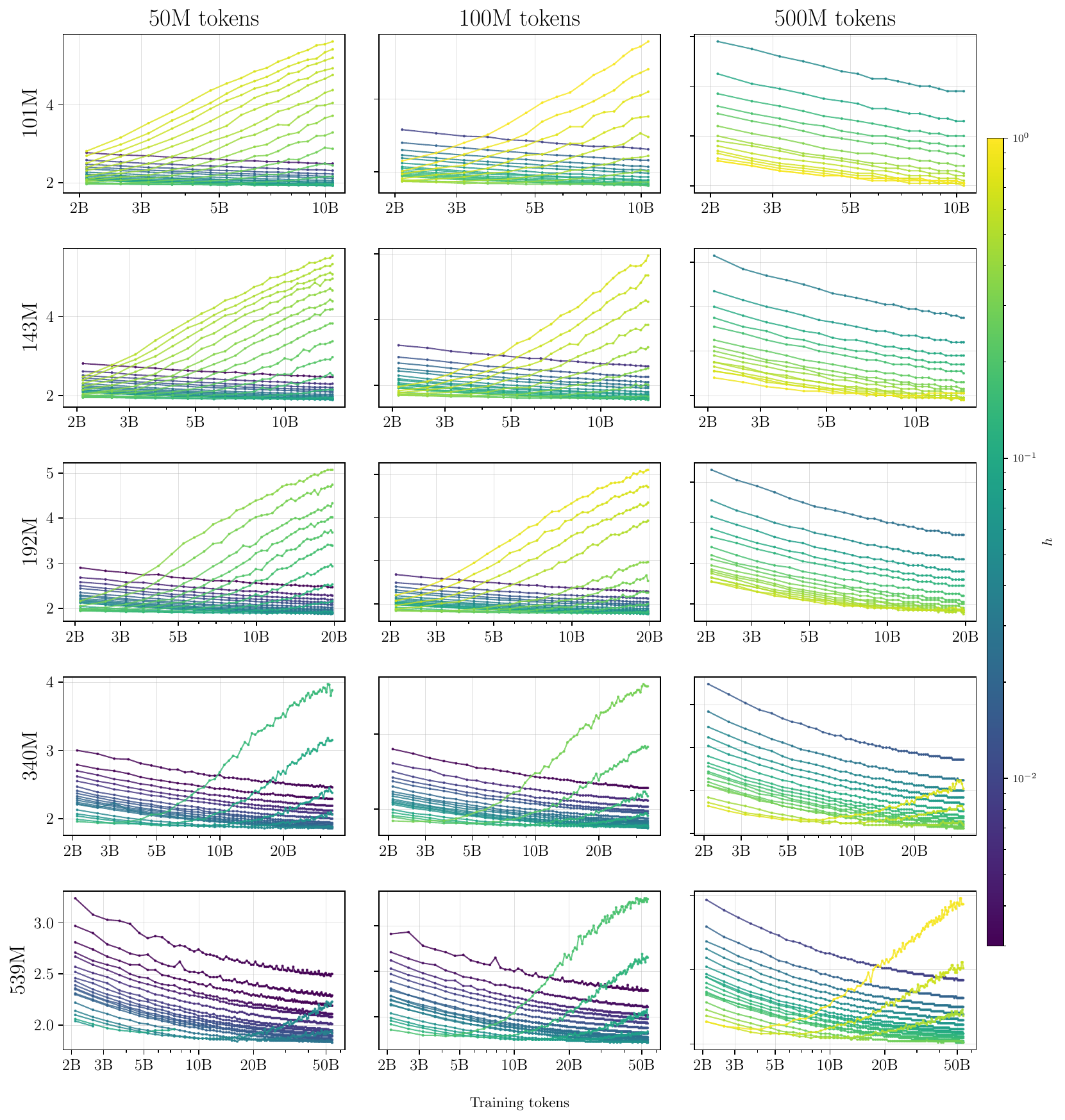}
\caption{Swahili target-domain loss curves across all model sizes (rows) and data pool sizes (columns).}
\label{fig:app_loss_sw}
\end{figure}

Figures~\ref{fig:app_loss_de}--\ref{fig:app_loss_sw} show target-domain validation loss as a function of training tokens for all model sizes (101M--539M) and all three target languages (German, French, Swahili). Each column corresponds to a different target data pool size; each row to a different model size; each curve within a panel corresponds to a different target weight $h$ (color). The same qualitative pattern(i.e.,  overfitting learning curves at high $h$ with small pools, while monotonic improvement with large pools) holds consistently across all configurations.

\begin{figure}[h!]
\centering
\includegraphics[width=\textwidth]{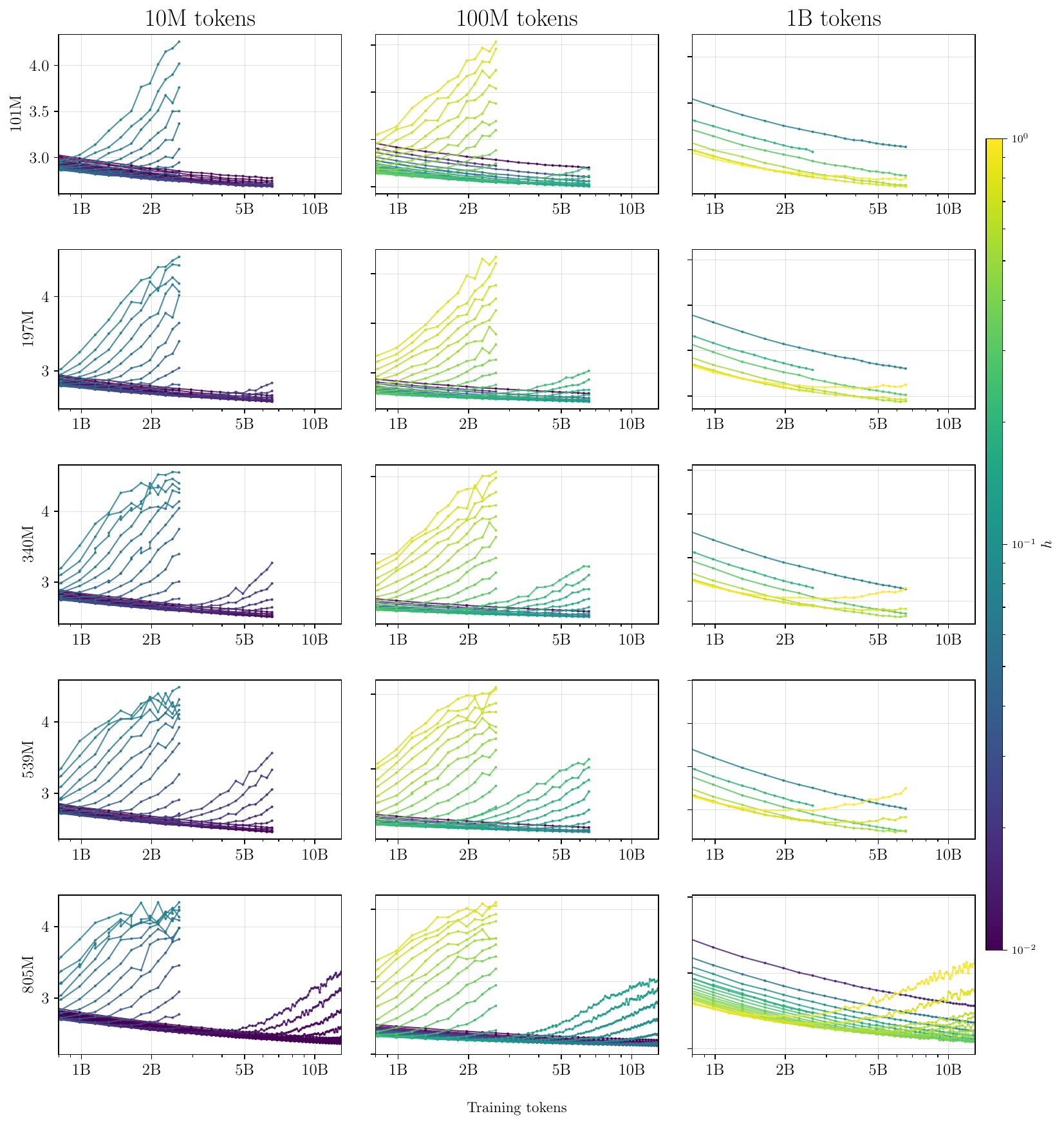}
\caption{Math (OpenWebMath) target-domain loss curves across all model sizes (rows) and data pool sizes (columns).}
\label{fig:app_loss_math}
\end{figure}

Figure~\ref{fig:app_loss_math} shows the same analysis for OpenWebMath, confirming that the loss curve dynamics extend beyond natural language to mathematical text.

\newpage

\newpage

\newpage

\newpage
\section{Cross-Model Loss Curves}
\label{app:cross_model}

\begin{figure}[h!]
\centering
\includegraphics[width=\textwidth]{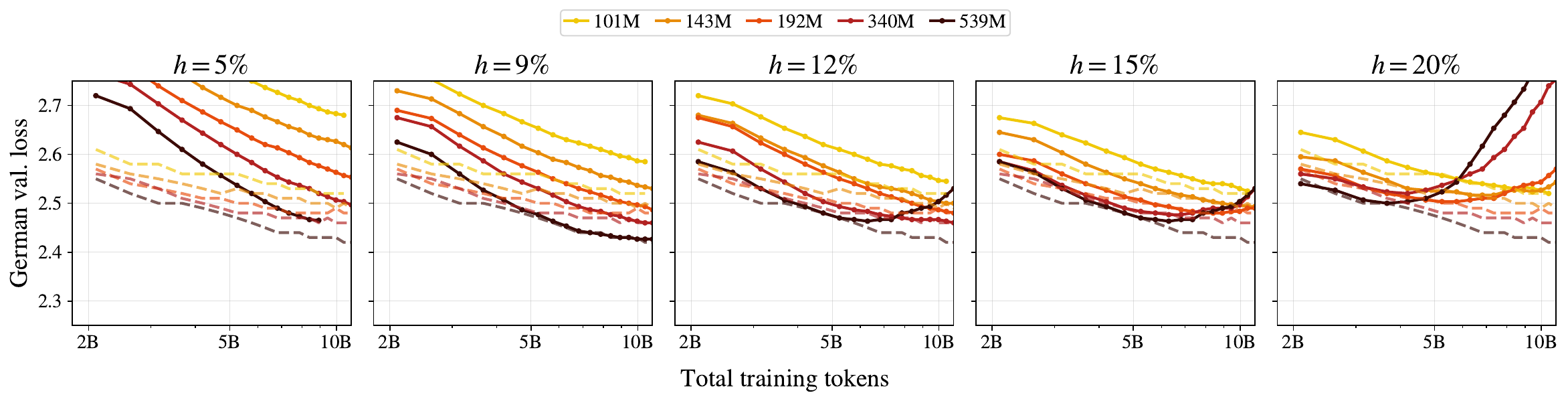}
\caption{German target-domain loss across model sizes at five target fractions, for the 100M token pool. Dashed lines show the best-loss envelope.}
\label{fig:cross_model_de}
\end{figure}

\begin{figure}[h!]
\centering
\includegraphics[width=\textwidth]{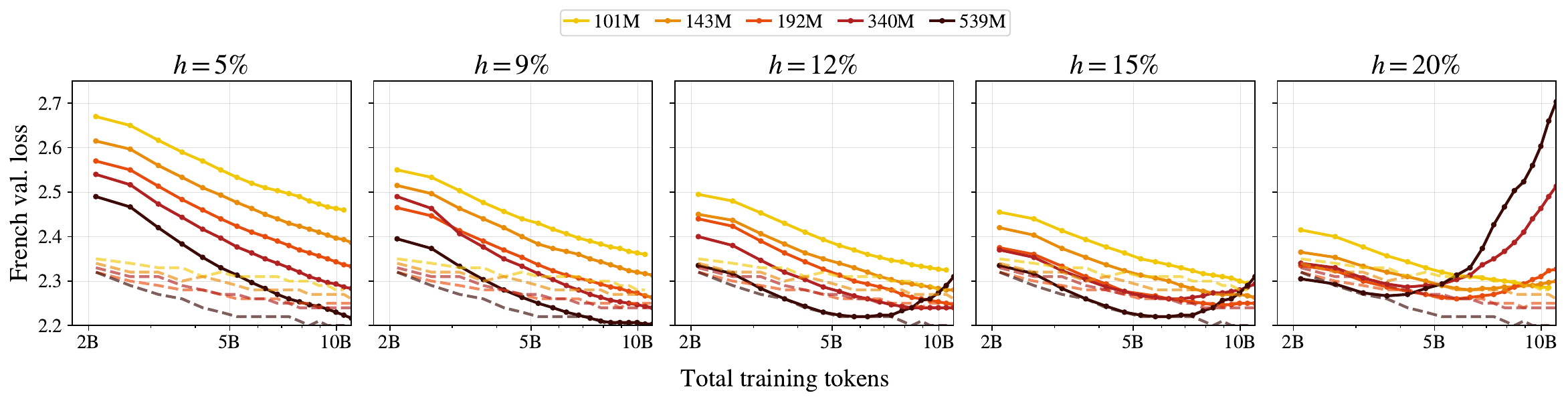}
\caption{French target-domain loss across model sizes at five target fractions, for the 100M token pool. Dashed lines show the best-loss envelope.}
\label{fig:cross_model_fr}
\end{figure}

\begin{figure}[h!]
\centering
\includegraphics[width=\textwidth]{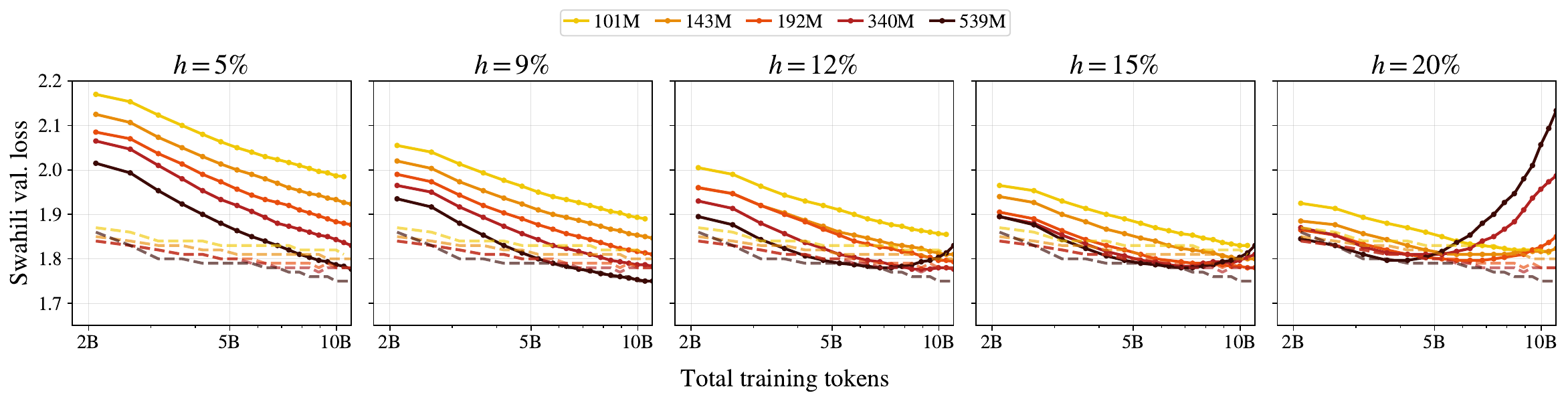}
\caption{Swahili target-domain loss across model sizes at five target fractions, for the 100M token pool. Dashed lines show the best-loss envelope.}
\label{fig:cross_model_sw}
\end{figure}

\begin{figure}[h!]
\centering
\includegraphics[width=\textwidth]{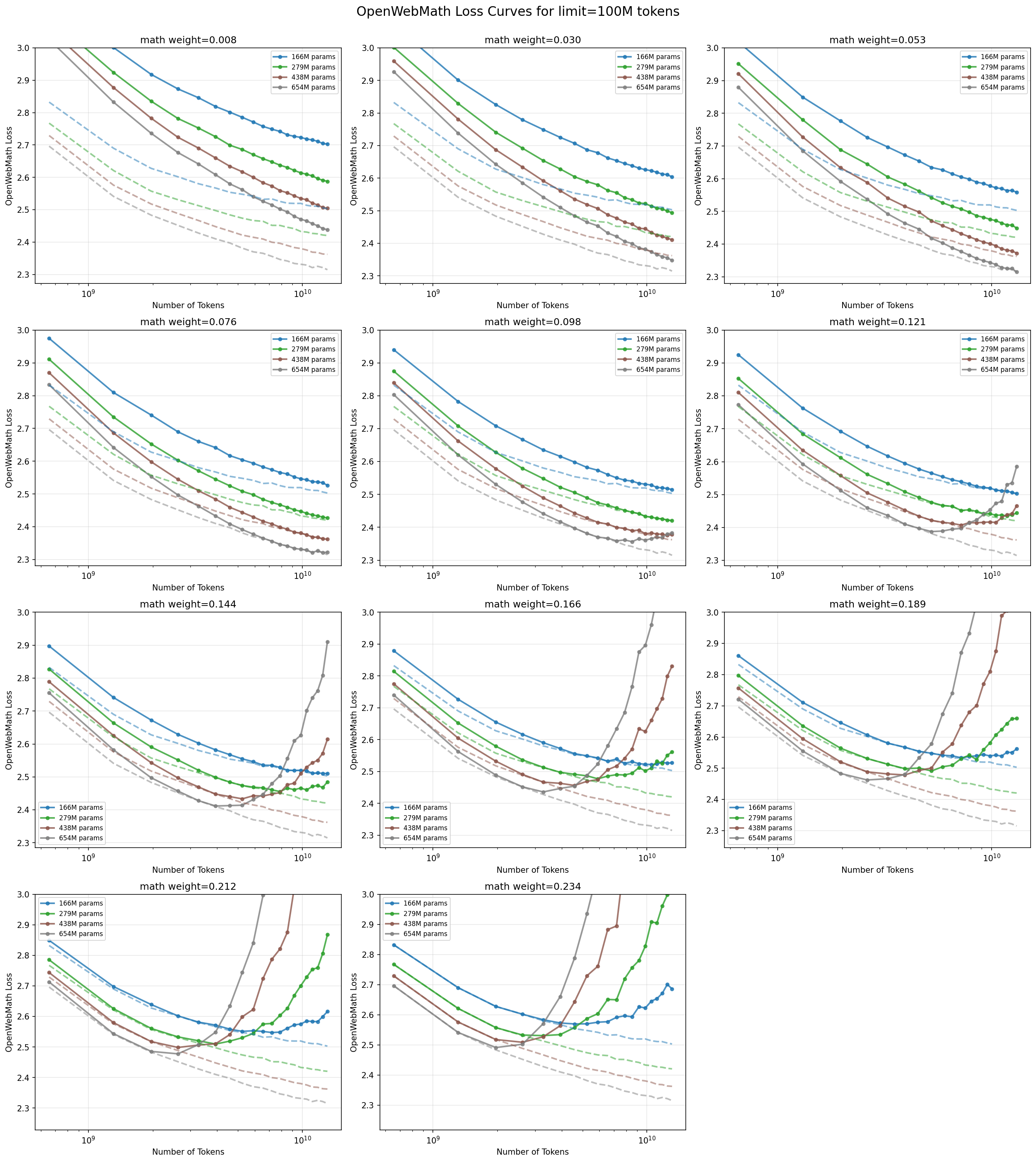}
\caption{OpenWebMath target loss across model sizes at different mixture fractions $h$, for the 100M token pool. Dashed lines show the best-loss envelope across all $h$ values.}
\label{fig:math_cross_model}
\end{figure}

Figure~\ref{fig:cross_model_fixed_h} shows that larger models overfit faster yet achieve lower best-case loss, demonstrated for German with 100M tokens at $h = 9\%$ and $20\%$. Here we extend this analysis to five target fractions (5--20\%) and verify the pattern across all languages and domains. Figure~\ref{fig:cross_model_de} shows the full German panel, Figures~\ref{fig:cross_model_fr}--\ref{fig:cross_model_sw} show French and Swahili, and Figure~\ref{fig:math_cross_model} shows the OpenWebMath domain. In all cases, the same ordering persists: larger models reach lower loss minima despite earlier overfitting onset, and the best-loss envelope (dashed) consistently favors larger models at every training budget. The pattern is particularly pronounced for the mathematics domain (Figure~\ref{fig:math_cross_model}), where the separation between model sizes is larger and the overfitting onset for larger models is sharper.

\newpage
\section{Quality Experiments: Additional Target Data Scales and Overlapping Quality Pools}
\label{app:quality_scales}

\begin{figure}[h!]
    \centering
    \includegraphics[width=\textwidth]{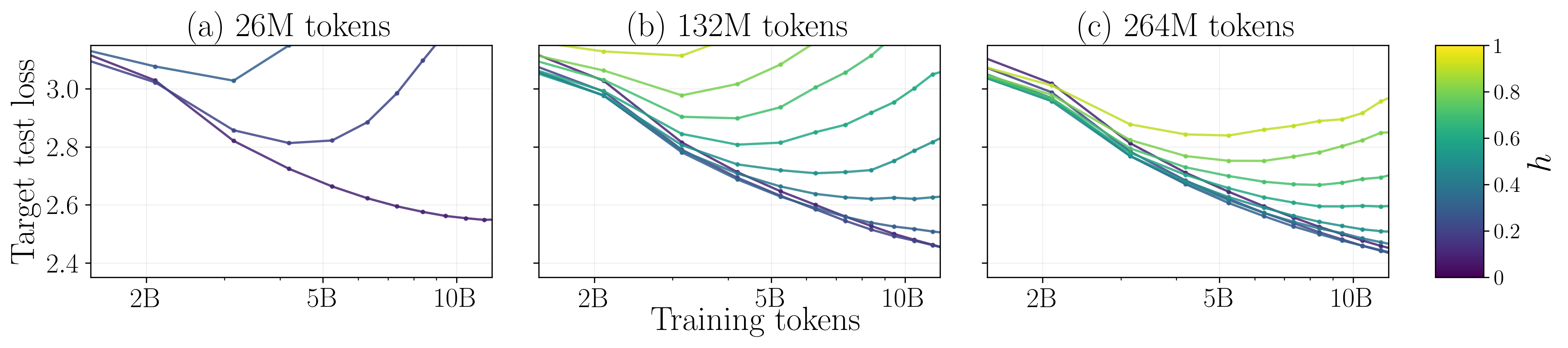}
    \caption{Loss vs. training tokens for Q99--100 at three data pool sizes. Each line corresponds to a target weight $h$. Smaller pools force lower weights to avoid overfitting from repetition.}
    \label{fig:top_1_data_constrained}
\end{figure}

To ground the experiments, we first confirm that a data constrained fixed quality level follows the same pattern as multilingual data. Figure~\ref{fig:top_1_data_constrained} shows Q99--100 loss curves at three pool sizes (26M, 132M, 264M tokens). At the smallest scale, only $h \leq 0.1$ avoids overfitting, and larger pools progressively unlock higher sustainable target weights.
This establishes the baseline that for a fixed quality threshold, the repetition penalty imposes the same constraint on the target loss.

\Cref{sec:results} further establishes that when training with a data-constrained high-quality domain, relaxing the quality filter to increase pool size consistently outperforms repeating the highest-quality slice starting from the 26M 99th percentile data pool.
A crossover at 5B tokens shows the optimal threshold shifts from Q95--99 to Q90--99 as training progresses. Here we present an expanded experimental suite, including additional data scales that support and extend these findings. We conduct three families of experiments, all using 100M-parameter models trained on a two-domain mixture of high-quality (HQ) filtered text and unlimited general DCLM. Each experiment sweeps target weight $w \in \{0.1, 0.2, \ldots, 1.0\}$.

\paragraph{Non-overlapping quality buckets at multiple data scales}
We define non-overlapping quality bands (Q90--99, Q95--99, Q99--100) and train at three data scales:
\begin{itemize}[nosep]
    \item \textbf{1$\times$}: pools range from 26M to 982M tokens including severe data constraint, and high repetition especially for high quality data.
    \item \textbf{5$\times$}: pools range from 132M to 2.1B tokens. This imposes a moderate constraint where the highest qualitybuckets have high repetition counts, but lower quality buckets do not.
    \item \textbf{10$\times$}: pools range from 264M to 4.2B tokens. Most quality buckets avoid significant repetition, and only the highest quality bucket still overfits with too much repetition.
\end{itemize}

\begin{figure}[h!]
  \centering
  \begin{subfigure}[t]{\textwidth}
      \centering
      \includegraphics[width=\textwidth]{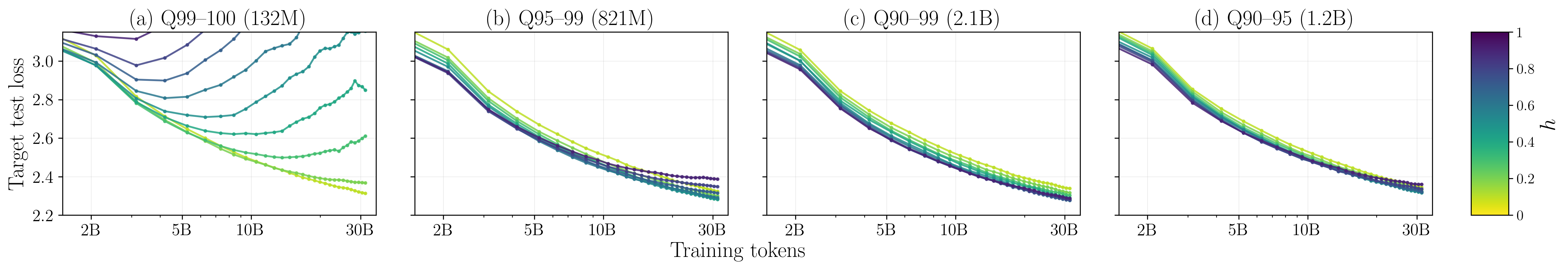}
      \caption{Pool sizes: Q99–100 (132M), Q95–99 (821M), Q90–99 (2.1B), Q90–95 (1.2B).}
      \label{fig:quality_1x}
  \end{subfigure}

  \begin{subfigure}[t]{\textwidth}
      \centering
      \includegraphics[width=\textwidth]{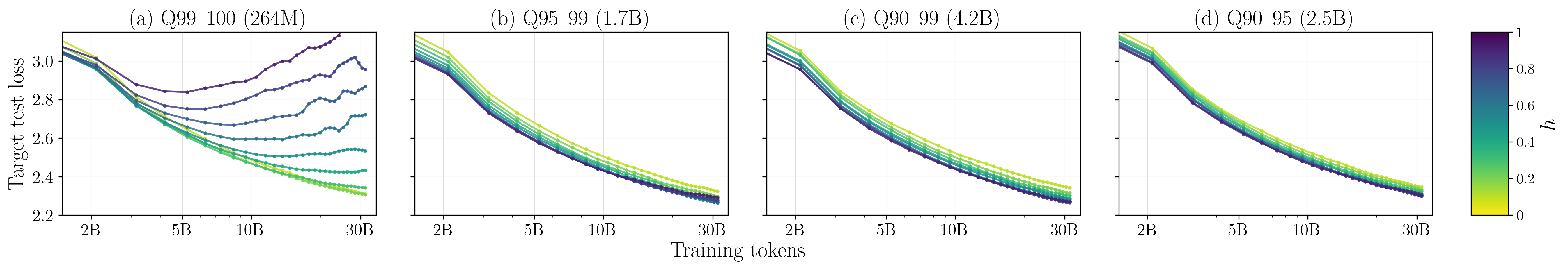}
      \caption{Pool sizes: Q99–100 (264M), Q95–99 (1.7B), Q90–99 (4.2B), Q90–95 (2.5B).}
      \label{fig:quality_2x}
  \end{subfigure}

  \caption{Loss curves for four quality bands at increasing data scales. Each line corresponds to a mixture weight $h$. Broader quality filters enable higher mixture weights without overfitting, with the effect diminishing as pool size increases.}
  \label{fig:quality_1x_2x}
  \end{figure}

\begin{figure}[h!]
  \centering
  \begin{subfigure}[t]{0.48\textwidth}
      \centering
      \includegraphics[width=\textwidth]{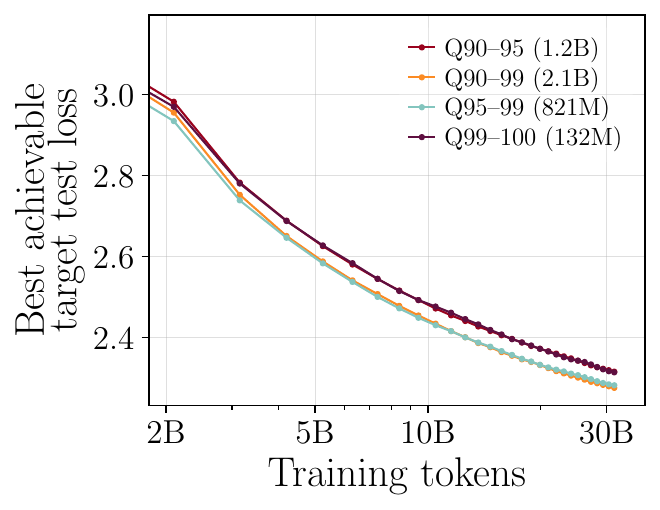}
      \caption{5$\times$ data scale.}
      \label{fig:quality_best_loss_1x}
  \end{subfigure}%
  \hfill
  \begin{subfigure}[t]{0.48\textwidth}
      \centering
      \includegraphics[width=\textwidth]{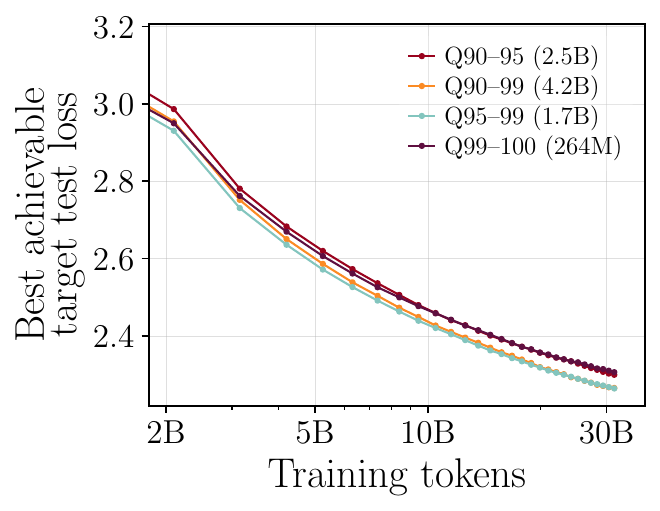}
      \caption{10$\times$ data scale.}
      \label{fig:quality_best_loss_2x}
  \end{subfigure}

  \caption{Best achievable target loss by quality band at (a) 5$\times$ and (b) 10$\times$ target data scales. Unlike the 1$\times$ setting (\cref{fig:comparison_best_loss}), at larger pool sizes no single quality band dominates: the curves largely overlap, confirming that the quality-quantity tradeoff diminishes when pools are large enough to avoid severe repetition.}
  \label{fig:quality_best_loss_1x_2x}
  \end{figure}

These experiments disentangle quality from quantity by ensuring no data overlap between bands, and reveal how the quality--quantity tradeoff evolves with different data constrained data scales.  Scaling up from the 26M target data pool to 5$\times$ and 10$\times$ the pool size confirms the main findings and reveals how the quality--quantity tradeoff evolves with data availability. At 5$\times$, Q95--99 dominates for the first ${\sim}$14B tokens before Q90--99 overtakes it (\Cref{fig:quality_1x}). The same crossover is observed at 1$\times$ but shifted later due to the larger pool delaying saturation. At 10$\times$, Q95--99 remains the best band throughout 30B tokens of training (\Cref{fig:quality_2x}): with data pools exceeding 1.6B tokens, no band experiences meaningful repetition at reasonable values of $h$, and higher per-token quality is the dominant factor. This demonstrates that the quality--quantity tradeoff is scale-dependent: at small scales, breadth wins; at large scales, quality wins; and the crossover token count increases predictably with pool size.  We include results with Q90-95 as a control to show that simply increasing the amount of data but decreasing quality performs worse.  The best-loss per quality over all $h$ is shown in \Cref{fig:quality_best_loss_1x_2x}, and confirms this quantitatively: at 1$\times$ starting from 132M target data size, a crossover from Q95--99 to Q90--99 occurs at ${\sim}$14B tokens (later than the ${\sim}$5B crossover at 1$\times$), while at 10$\times$ Q95--99 leads throughout. As pool size grows, the quality--quantity tradeoff vanishes and per-token quality is most important.

\begin{figure}[h!]
\centering
\includegraphics[width=\textwidth]{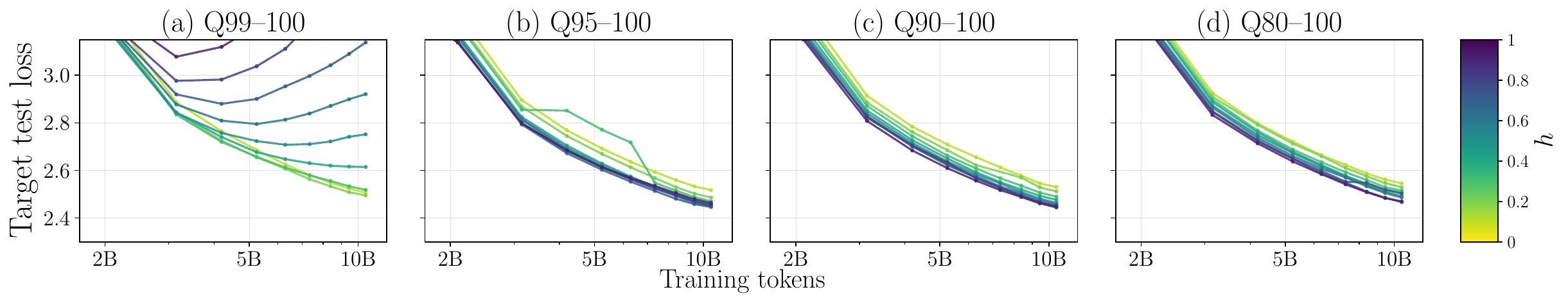}
\caption{Loss curves for inclusive quality thresholds (each band includes all data above the given percentile). Pool sizes increase from left to right as broader thresholds include more data.}
\label{fig:quality_inclusive}
\end{figure}

\paragraph{Experiments with overlapping buckets}
We use the full untruncated pool at each quality threshold (Q80--100: 5B, Q90--100: 2B, Q95--100: 954M, Q99--100: 132M) and train for 10B tokens.  In this expeirment, the quality bands are (e.g., all contain the Q99-100 data), representing the practical approach of setting a single lower-bound quality threshold.  The results are consistent.  The 95th percentile, Q95--100, achieves the best loss consistent with the Q95-99 results for up to 10B tokens in \Cref{fig:quality_inclusive}.

\begin{figure}[h!]
\centering
\includegraphics[width=0.45\textwidth]{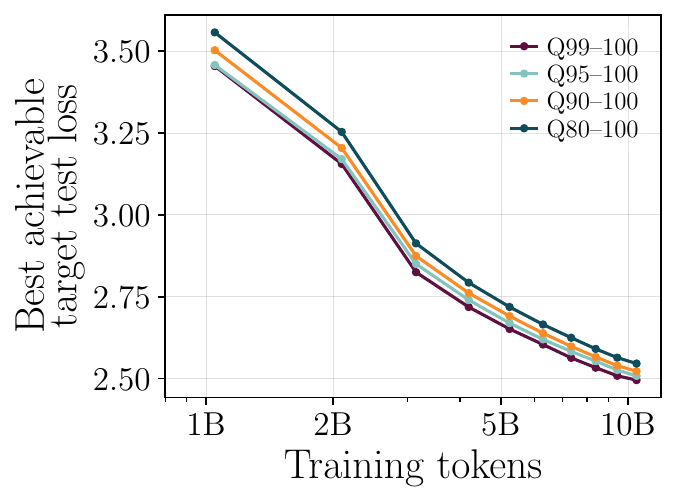}
\caption{Best achievable target loss with all quality bands truncated to the same pool size. With repetition equalized, higher per-token quality consistently achieves lower loss.}
\label{fig:quality_same_pool}
\end{figure}

\paragraph{Equal-size data constrained size experiments.}
We truncate each quality band (Q80--100, Q90--100, Q95--100, Q99--100) to approximately the same pool size (${\sim}$132--139M tokens) and train for 10B tokens. This isolates the effect of per-token quality by controlling for pool size as all bands have approximately the same number of repetitions. \ref{fig:quality_same_pool} shows results when the data constrained pools are equal size.  With the impact of repetition equalized across quality, the ordering is strict: Q99--100 $>$ Q95--100 $>$ Q90--100 $>$ Q80--100 at all training steps.

\paragraph{Final remarks} When pool size is equalized, quality improves validation loss.  When pools vary naturally, the repetition penalty from small high-quality pools outweighs the quality advantage. The non-overlapping band experiments isolate the mechanism: it is the data in broader bands (Q90--99 beyond Q95--99) that helps at larger training budgets, not merely reduced repetition of the top slice. The scale dependence of the crossover has a practical implication that for any given training budget and data pool, there exists an optimal quality threshold that balances per-token quality against repetition.  Our scaling law captures this tradeoff enabling practitioners to extrapolate which data pool to use and for how many repetitions.

\section{Comparison to Unlimited Target Data}
\label{app:unlimited}

\begin{figure}[h!]
\centering
\includegraphics[width=\textwidth]{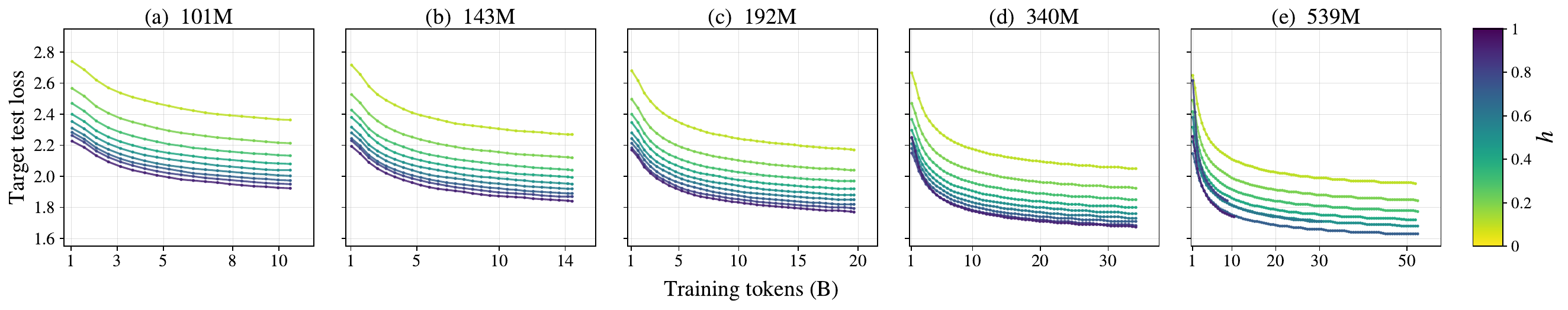}
\caption{German target loss with unlimited data (${\sim}60$B pool, $r < 1$ throughout) for all mixture fractions $h$. Without data constraints, higher target fractions always achieve lower target loss, and no overfitting occurs.}
\label{fig:unlimited_all_h}
\end{figure}

To establish an upper bound on target-domain performance, we train all model sizes with a ${\sim}60$B German token pool -- large enough that repetition never occurs. Figure~\ref{fig:unlimited_all_h} shows target loss for each mixture fraction $h$ in this unlimited setting. Without data constraints, higher target fractions monotonically achieve lower target loss at every training budget: there is no overfitting, no U-shaped behavior, and no quality--quantity tradeoff. 

\begin{figure}[h!]
\centering
\includegraphics[width=\textwidth]{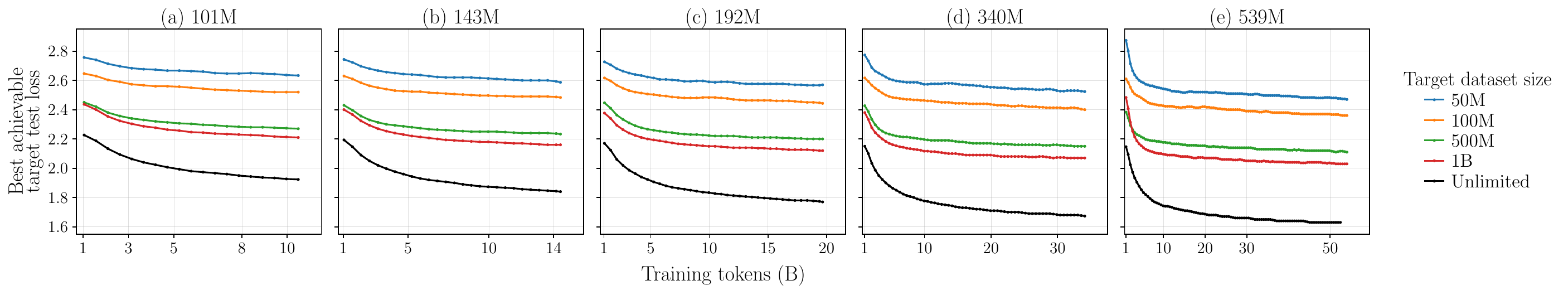}
\caption{Best achievable German target loss vs.\ training tokens for each model size, comparing constrained data pools against an unlimited (${\sim}60$B) baseline where target data is never repeated.}
\label{fig:unlimited_comparison}
\end{figure}

Figure~\ref{fig:unlimited_comparison} then compares the best achievable loss under constrained pools (50M--1B) against this unlimited baseline. The gap between each constrained curve and the unlimited line quantifies the cost of the data constraint. Constrained pools plateau as training progresses while the unlimited baseline continues to improve, confirming that the performance ceiling is imposed by data exhaustion rather than model capacity. The gap widens with training budget, underscoring the importance of optimal mixture design when target data is scarce.

\section{Analysis of Three-Domain Experiments}
\label{app:3domain}

Section~\ref{sec:multi_domain} presents results under proportional weighting; here we report the complementary equal-weighting experiments ($h_{\text{wiki}} = h_{\text{peS2o}}$).

\begin{figure}[h!]
\centering
\vspace{-10pt}
\includegraphics[width=0.45\textwidth]{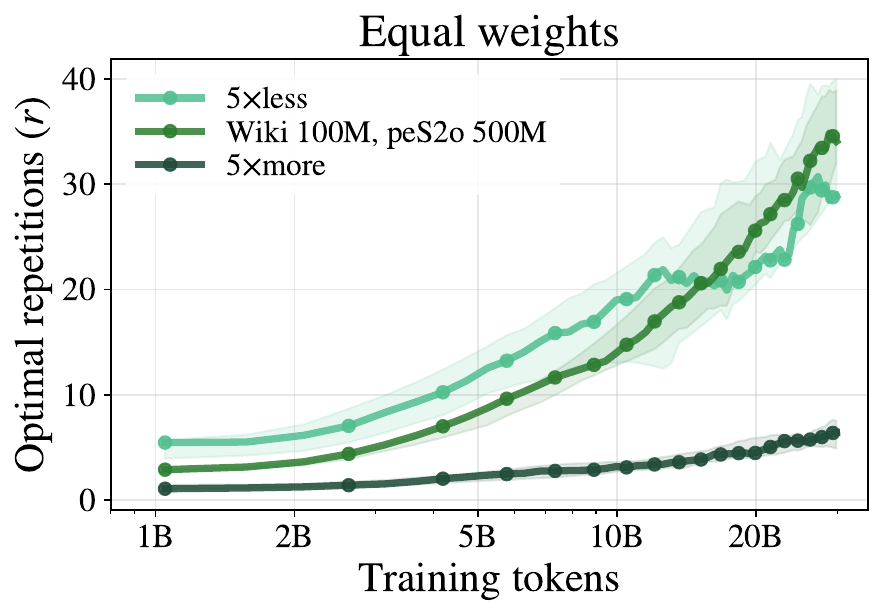}
\caption{Optimal repetitions under equal weighting with 10\% compute confidence band (shaded) in the three-domain setup (101M model).}
\label{fig:3domain_equal}
\end{figure}

Figure~\ref{fig:3domain_equal} shows optimal repetition under equal weighting with the 10\% compute confidence band. The core pattern from the proportional case carries over: optimal $r$ grows steadily with training budget, and smaller data pools require substantially higher repetition before the optimum is reached. Under equal weighting, the smaller domain (Wiki) is repeated more aggressively than under proportional weighting, as equal weights allocate the same weight to both domains regardless of pool size, and the smaller pool exhausts its unique tokens earlier. For example, at 10B training tokens in the base configuration (Wiki 100M, peS2o 500M), equal weighting reaches $r \approx 14$ while proportional weighting stays at $r \approx 6$ (measured with respect to the Wiki pool). The confidence bands remain wide, confirming that approximate repetition estimates suffice for near-optimal performance under either strategy.

\begin{figure}[h!]
    \centering
    \includegraphics[width=0.55\textwidth]{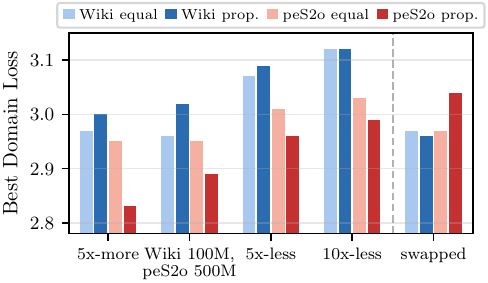}
    \caption{Per-domain best loss under equal vs.\ proportional weighting across five pool-size configurations.}
    \label{fig:3domain_bars}
    \vspace{-10pt}
\end{figure}

Additionally, we examined how the choice of weighting strategy affects each target domain individually. Beyond the configurations reported in the main text, we include a 10$\times$less experiment (Wiki 10M, peS2o 50M) to test both strategies under even more severe data constraints. Figure~\ref{fig:3domain_bars} breaks down the per-domain loss under both strategies across all five configurations. Proportional weighting consistently favors the larger pool (peS2o), achieving lower peS2o loss at the expense of Wiki. Equal weighting reverses this tradeoff: it assigns more relative weight to the smaller domain, improving Wiki loss but sacrificing peS2o performance. The effect is most pronounced when pool sizes differ substantially: in the 5$\times$more configuration, proportional weighting achieves 0.12 lower peS2o loss than equal weighting, while equal weighting gains only 0.03 on Wiki. The \emph{swapped} configuration (where Wiki becomes the larger pool) confirms this pattern: when pool sizes are reversed, proportional weighting now favors Wiki instead.

\begin{figure}[h!]
\centering
\includegraphics[width=0.65\textwidth]{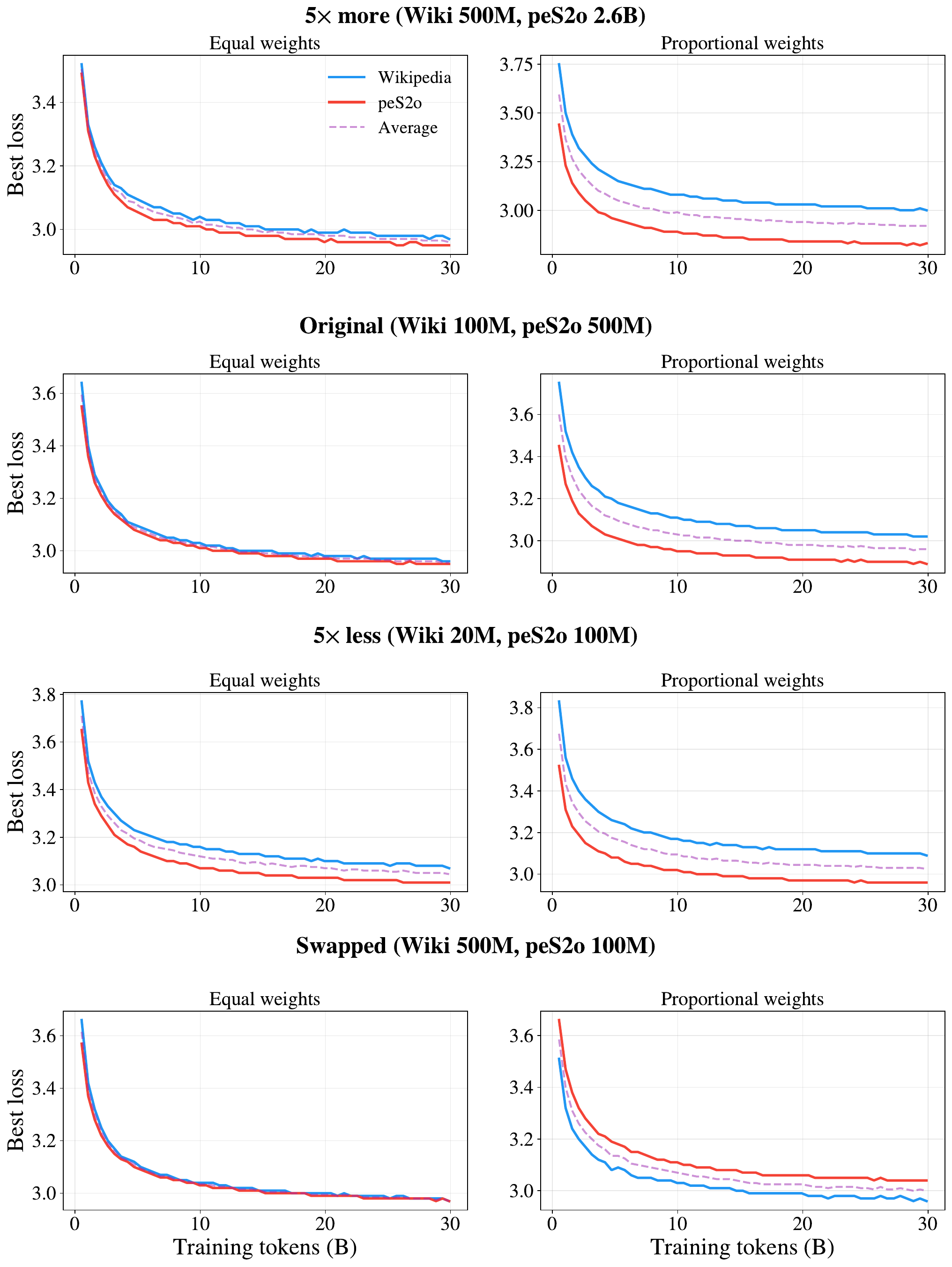}
\caption{Best achievable per-domain loss throughout training under equal (left) and proportional (right) weighting for four pool-size configurations. Blue: Wikipedia; red: peS2o; dashed purple: average.}
\vspace{-5pt}
\label{fig:3domain_optimal_loss_all}
\end{figure}

Figure~\ref{fig:3domain_optimal_loss_all} shows the best achievable per-domain loss throughout training for all four pool-size configurations.

\newpage
\section{Baseline Scaling Law Formulas}
\label{app:details}

We describe the baseline formulas used in Section~\ref{sec:scaling_results}.
All formulas predict the target-domain loss as a function of the total tokens $D_\text{total}$, the target weight $h$, the number of repetitions $r = h\,D_\text{total}/D_\text{target}$, and optionally the model size $N$.

\paragraph{Repetitions-agnostic~\citep{shukor2025scalinglawsoptimaldata}.}
This baseline treats all tokens as unique regardless of repetition:
\begin{equation}
  L = E + \frac{A}{D_\mathrm{eff}^\alpha} + \gamma\,h\,,
  \qquad D_\mathrm{eff} = (1-h)\,D_\text{total} + \tau\,h\,D_\text{total}\,.
\end{equation}
The effective data is a weighted sum of generic and target tokens with no repetition discounting---repeated target tokens are valued the same as fresh ones.
Parameters: $E$, $A$, $\alpha$, $\tau$, $\gamma$ (5 parameters).

\paragraph{Domain-agnostic~\citep{muennighoff2023scaling}.}
This baseline uses a single saturating function of total tokens without distinguishing domains:
\begin{equation}
  L = E + A\,D_\mathrm{eff}^{\,\alpha}\,,
  \qquad D_\mathrm{eff} = C\bigl(1 - e^{-\mu\,R}\bigr)\,,
\end{equation}
where $C = (1-h)\,D_\text{total} + h\,D_\text{total}/r$ is the total number of unique tokens and $R = D_\text{total}/C$ is the overall repetition ratio.
Parameters: $E$, $A$, $\alpha$ ($<0$), $\mu$ (4 parameters).

\paragraph{Utility decay~\citep{goyal2024scaling}.}
This baseline models repetition through an exponentially decaying exponent:
\begin{equation}
  L = E + a\,D_\text{total}^{\,b_\mathrm{eff}}\,,
  \qquad b_\mathrm{eff} = (1-h)\,b_0 + h\,b_1\,\delta^{r-1}\,,
\end{equation}
where $\delta = 0.5^{1/\tau}$ is a half-life parameter.
As repetitions increase, the exponent on the target-domain contribution decays toward zero, reducing the data-scaling benefit.
Parameters: $E$, $a$, $b_0$, $b_1$, $\tau$ (5 parameters).

\paragraph{Multi-size extensions.}
Each baseline extends to variable model size by adding a capacity term $C/N^\beta$ (and for the repetitions-agnostic and domain-agnostic formulas, a data-size coupling $N^\delta$), following the same pattern as $L_\mathrm{size}$ (Eq.~\ref{eq:lsize}).

\section{Extended Scaling Law Results}
\label{app:extended_scaling}

This section reports the full results underlying Section~\ref{sec:scaling_results}, including the train/test breakdown for the loss prediction tables, the per-dataset optimal-mixture prediction errors, and the size-extrapolation wasted-token metric.

\paragraph{Loss prediction: train and test $wR^2$ (fixed-size formula).}
Table~\ref{tab:app_scaling_r2_full} reports train and test $wR^2$ on all four experimental setups for the fixed-size formula $L_\mathrm{fix}$ and the three baselines. Train corresponds to the first 50\% of training steps, test to the held-out second half.

\begin{table}[h]
\centering
\caption{Train and test weighted $R^2$ ($wR^2$) for the fixed-size formula $L_\mathrm{fix}$, fitted independently per model size (or per quality level / domain). Each observation is weighted by $\max(r \cdot h, \epsilon)$ to emphasise the high-repetition regime. Train: first 50\% of steps; test: second half.}
\label{tab:app_scaling_r2_full}
\small
\begin{tabular}{lcccccccc}
\toprule
& \multicolumn{2}{c}{German} & \multicolumn{2}{c}{Maths} & \multicolumn{2}{c}{Quality} & \multicolumn{2}{c}{Wiki/peS2o} \\
\cmidrule(lr){2-3} \cmidrule(lr){4-5} \cmidrule(lr){6-7} \cmidrule(lr){8-9}
& Train & Test & Train & Test & Train & Test & Train & Test \\
\midrule
$L_\mathrm{fix}$ & \textbf{0.85} & \textbf{0.95} & \textbf{0.99} & \textbf{0.88} & \textbf{0.91} & \textbf{0.71} & 0.68 & \textbf{0.80} \\
Rep-agnostic & 0.81 & 0.78 & 0.95 & 0.78 & 0.86 & 0.14 & 0.67 & 0.72 \\
Utility decay & 0.52 & 0.72 & 0.77 & 0.55 & 0.94 & $-$0.64 & \textbf{0.69} & 0.79 \\
Domain-agnostic & 0.03 & $-$40.7 & 0.10 & $-$0.49 & 0.46 & $-$2.19 & 0.42 & $-$1.16 \\
\bottomrule
\end{tabular}
\end{table}

\paragraph{Loss prediction: train and test $wR^2$ (size extrapolation).}
Table~\ref{tab:app_scaling_r2_size_full} reports the same metric for the multi-size formula $L_\mathrm{size}$, fitted on all model sizes except the largest and evaluated on the held-out largest model.

\begin{table}[h]
\centering
\caption{Size extrapolation: weighted $R^2$ ($wR^2$) for the multi-size formula $L_\mathrm{size}$, fitted on smaller model sizes and evaluated on the held-out largest model (539M for German, 936M for Maths). Train: $wR^2$ on fitting sizes; Test: $wR^2$ on held-out size.}
\label{tab:app_scaling_r2_size_full}
\small
\begin{tabular}{lcccccc}
\toprule
& \multicolumn{2}{c}{German (539M)} & \multicolumn{2}{c}{Maths (936M)} \\
\cmidrule(lr){2-3} \cmidrule(lr){4-5}
& Train & Test & Train & Test \\
\midrule
$L_\mathrm{size}$ & \textbf{0.95} & \textbf{0.65} & \textbf{0.97} & \textbf{0.73} \\
Rep-agnostic+$N$ & 0.88 & 0.59 & 0.94 & 0.71 \\
Utility decay+$N$ & 0.76 & 0.56 & 0.85 & 0.69 \\
Domain-agnostic+$N$ & 0.25 & $-$0.23 & 0.46 & $-$0.77 \\
\bottomrule
\end{tabular}
\end{table}

\paragraph{Optimal mixture prediction error.}
For each experiment (a specific model size and target dataset size) and each held-out training checkpoint $D_\text{total}$, we compare the predicted $h^*$ to the empirically best weight $h^*_\text{emp}$ at that checkpoint. Table~\ref{tab:app_hstar_error} reports the median of $|\log_{10}(h^*_\text{pred}) - \log_{10}(h^*_\text{emp})|$ per dataset over all such pairs on held-out test steps. A value of $0.1$ corresponds to a ${\sim}1.26\times$ multiplicative error in the predicted weight.

\begin{table}[h]
\centering
\caption{Optimal mixture prediction on held-out test steps: median $|\log_{10}(h^*_\text{pred}) - \log_{10}(h^*_\text{emp})|$ (lower is better). A value of $0.1$ corresponds to a ${\sim}1.26\times$ error in the predicted fraction.}
\label{tab:app_hstar_error}
\small
\begin{tabular}{lccccc}
\toprule
& Ger. & Math & Qual. & Wiki & Avg \\
\midrule
$L_\mathrm{fix}$ & \textbf{0.07} & \textbf{0.13} & 0.47 & \textbf{0.10} & \textbf{0.19} \\
Rep-agnostic & 0.60 & 1.01 & 0.61 & 0.50 & 0.68 \\
Utility decay & 0.10 & 0.20 & \textbf{0.38} & 0.12 & 0.20 \\
Domain-agn. & 2.57 & 1.99 & 5.30 & 3.07 & 3.23 \\
\midrule
$L_\mathrm{size}$ & \textbf{0.15} & 0.18 & --- & --- & 0.17 \\
Rep-agn.+$N$ & 0.73 & 1.44 & --- & --- & 1.09 \\
Util. dec.+$N$ & \textbf{0.15} & \textbf{0.17} & --- & --- & \textbf{0.16} \\
Dom.-agn.+$N$ & 3.27 & 1.56 & --- & --- & 2.42 \\
\bottomrule
\end{tabular}
\end{table}

\paragraph{Size extrapolation: wasted tokens.}
A stronger test of the multi-size formula is whether it can predict the behaviour of a model size never seen during fitting. We fit $L_\mathrm{size}$ and all multi-size baselines on all model sizes \emph{except the largest} and evaluate exclusively on the held-out largest model (539M for German, 936M for Maths). Table~\ref{tab:size_extrap_wasted} reports the weighted median wasted token fraction on the held-out largest model for each setup.
$L_\mathrm{size}$ achieves a test $wR^2$ of $0.65$ on German and $0.73$ on Maths (Table~\ref{tab:app_scaling_r2_size_full}), demonstrating that the capacity and data-size coupling terms extrapolate meaningfully to larger scales.
The wasted token metric is even more discriminating: most baselines fail catastrophically on German, producing predictions outside the empirical range (counted as~$100\%$).
$L_\mathrm{size}$ wastes $58\%$ on German and $24\%$ on Maths, the only formula that avoids catastrophic failure on both.
Results for French and Swahili are reported in Appendix~\ref{app:size_extrap}.

\begin{table}[h]
\centering
\caption{Size extrapolation: weighted median wasted token fraction (lower is better). Predictions outside the empirical weight range are counted as $100\%$.}
\label{tab:size_extrap_wasted}
\small
\begin{tabular}{lcc}
\toprule
& German (539M) & Maths (936M) \\
\midrule
$L_\mathrm{size}$ & \textbf{58\%} & \textbf{24\%} \\
Rep-agnostic+$N$ & 100\% & 97\% \\
Utility decay+$N$ & 100\% & 22\% \\
Domain-agnostic+$N$ & 100\% & 84\% \\
\bottomrule
\end{tabular}
\end{table}

\paragraph{Limitations}
The scaling law is descriptive: it summarises observed training dynamics but does not prescribe an optimal allocation from first principles.
Size extrapolation is encouraging on both German ($wR^2 = 0.65$ at 539M) and Maths ($wR^2 = 0.73$ at 936M), though the wasted-token metric reveals that predictions remain imperfect at unseen scales.
The quality-filtering setup is harder to predict than language/domain splits (test $wR^2 = 0.71$, log-error $= 0.47$), likely because quality tiers interact with the model in ways not fully captured by the effective data abstraction.
Finally, our experiments span ${\sim}100\text{M}$--$900\text{M}$ parameters; whether the fitted exponents extrapolate to multi-billion-parameter models remains an open question.

\paragraph{Size Extrapolation: French and Swahili}
\label{app:size_extrap}

We repeat the size-extrapolation experiment of Section~\ref{sec:scaling_results} on French and Swahili.
For each language, we fit the multi-size formulas on model sizes 101M--340M and evaluate on the held-out 539M model.
Table~\ref{tab:size_extrap_app} reports weighted $R^2$ (test\,/\,train).

On French, $L_\mathrm{size}$ achieves a test $wR^2$ of $0.93$, even exceeding its train $wR^2$ of $0.92$, what indicates that the 539M model's behaviour is well predicted by trends from smaller scales.
On Swahili, extrapolation is harder: $L_\mathrm{size}$ achieves $0.69$ on the held-out 539M model, while the utility decay baseline reaches $0.71$.
The domain-agnostic baseline yields near-zero or negative $wR^2$ on both languages.

For wasted tokens, all formulas (including $L_\mathrm{size}$) produce predictions outside the empirical weight range for French and Swahili at 539M, resulting in ${\geq}94\%$ waste.
This reflects the narrower range of target fractions tested at this scale for these languages rather than a fundamental failure of the formula.

\begin{table}[h!]
\centering
\caption{Size extrapolation on French and Swahili: weighted $R^2$ (test\,/\,train) when fitting on 101M--340M and evaluating on 539M.}
\label{tab:size_extrap_app}
\begin{tabular}{lcc}
\toprule
& French (539M) & Swahili (539M) \\
\midrule
$L_\mathrm{size}$ (9p) & \textbf{0.93}\,/\,\textbf{0.92} & 0.69\,/\,\textbf{0.92} \\
Rep-agnostic+$N$ (8p) & 0.87\,/\,0.90 & 0.65\,/\,0.90 \\
Utility decay+$N$ (7p) & 0.69\,/\,0.62 & \textbf{0.71}\,/\,0.69 \\
Domain-agnostic+$N$ (7p) & $-$0.54\,/\,0.08 & $-$0.00\,/\,0.02 \\
\bottomrule
\end{tabular}
\end{table}

\section{Robustness to Weighting Scheme}
\label{app:weighting}

We verify that the qualitative ranking of formulas is robust to the choice of observation weights in the $wR^2$ metric.
Table~\ref{tab:weighting_robustness} reports test $wR^2$ on the German dataset under four weighting schemes: $\max(r \cdot h,\,\epsilon)$ (our default), $r$ only, $h$ only, and uniform (standard $R^2$).
The ranking $L_\mathrm{fix}$ $>$ rep-agnostic $>$ utility decay $>$ domain-agnostic is preserved across all schemes, confirming that the advantage of our formula is not an artifact of the weighting.
\begin{table}[h!]
\centering
\caption{Test $wR^2$ on German under different weighting schemes. The formula ranking is stable across all choices.}
\label{tab:weighting_robustness}
\begin{tabular}{lcccc}
\toprule
& $r \cdot h$ (default) & $r$ & $h$ & Uniform \\
\midrule
$L_\mathrm{fix}$ (6p) & \textbf{0.96} & \textbf{0.95} & \textbf{0.78} & \textbf{0.90} \\
Rep-agnostic (5p) & 0.81 & 0.82 & 0.52 & 0.85 \\
Utility decay (5p) & 0.78 & 0.85 & 0.39 & 0.71 \\
Domain-agnostic (4p) & $<$$-$1 & $<$$-$1 & $<$$-$1 & $<$$-$1 \\
\bottomrule
\end{tabular}
\end{table}

\section{Downstream Benchmark Evaluation}
\label{app:downstream}

We evaluate downstream task performance for the best-loss configurations from Figure~\ref{fig:diminishing} (one checkpoint per model size and data pool combination, selected at the training step that minimizes German test loss). We use seven benchmarks covering reasoning, commonsense, and language understanding:
ARC-Challenge and ARC-Easy \citep{clark2018think} (science question answering),
HellaSwag \citep{zellers2019hellaswag} (commonsense sentence completion),
LAMBADA \citep{paperno2016lambada} (long-range word prediction),
PIQA \citep{bisk2020piqa} (physical intuition reasoning),
SciQ \citep{welbl2017sciq} (science multiple-choice),
and WinoGrande \citep{sakaguchi2020winogrande} (coreference resolution).
Each benchmark is evaluated in both English (original) and German (translated), allowing us to assess both target-domain and non-target-domain capabilities from the same checkpoint.

Tables~\ref{tab:downstream_de} and~\ref{tab:downstream_en} report accuracy on all benchmarks.
German benchmark performance improves consistently with both model size and target data pool size, confirming that the perplexity gains translate to downstream improvements. English performance decreases as more training budget is allocated to German (higher $h$), reflecting the expected trade-off between target and non-target domains. Larger models partially mitigate this cost: the 539M model retains 48.6\% English average even at the highest German allocation, compared to 39.7\% for 101M.

\begin{table}[h]
\centering
\caption{German downstream benchmark accuracy (\%) at the best target-loss checkpoint. More target data consistently improves German performance across all model sizes.}
\label{tab:downstream_de}
\begin{tabular}{ll ccccccc c}
\toprule
Model & Dataset size & ARC-C & ARC-E & HSwag & Lambada & PIQA & SciQ & Wino & \textbf{Avg} \\
\midrule
101M & 50M  & 24.0 & 31.0 & 28.0 & 13.7 & 52.1 & 60.4 & 50.8 & 37.2 \\
101M & 100M & 24.5 & 31.2 & 28.4 & 15.2 & 53.0 & 63.2 & 50.2 & 37.9 \\
101M & 500M & 23.4 & 32.6 & 29.8 & 17.6 & 55.3 & 60.1 & 50.3 & 38.5 \\
101M & 1B   & 25.7 & 30.9 & 29.8 & 19.1 & 57.0 & 59.1 & 48.9 & 38.6 \\
\midrule
143M & 50M  & 23.2 & 29.9 & 28.6 & 15.3 & 52.4 & 61.7 & 50.1 & 37.3 \\
143M & 100M & 23.7 & 31.9 & 28.7 & 16.3 & 54.4 & 62.3 & 51.5 & 38.4 \\
143M & 500M & 23.7 & 33.5 & 29.8 & 19.3 & 56.5 & 63.2 & 50.8 & 39.6 \\
143M & 1B   & 24.3 & 33.1 & 30.5 & 20.7 & 56.0 & 62.8 & 50.7 & 39.7 \\
\midrule
192M & 50M  & 24.1 & 31.0 & 28.9 & 15.4 & 53.7 & 62.1 & 49.7 & 37.8 \\
192M & 100M & 24.2 & 32.7 & 29.6 & 16.9 & 53.9 & 63.2 & 50.5 & 38.7 \\
192M & 500M & 24.1 & 32.3 & 31.2 & 20.0 & 57.6 & 62.7 & 51.3 & 39.9 \\
192M & 1B   & 24.2 & 35.3 & 31.2 & 20.8 & 57.5 & 62.2 & 52.9 & 40.6 \\
\midrule
340M & 50M  & 24.2 & 31.2 & 29.2 & 15.7 & 52.8 & 62.6 & 51.4 & 38.2 \\
340M & 100M & 25.0 & 33.0 & 30.4 & 18.4 & 53.7 & 64.2 & 49.7 & 39.2 \\
340M & 500M & 24.8 & 33.8 & 32.2 & 24.1 & 57.6 & 65.3 & 53.0 & 41.5 \\
340M & 1B   & 25.3 & 36.1 & 32.9 & 24.6 & 57.2 & 64.3 & 52.5 & 41.8 \\
\midrule
539M & 50M  & 24.2 & 31.0 & 29.8 & 19.0 & 52.9 & 64.6 & 50.0 & 38.8 \\
539M & 100M & 26.1 & 31.8 & 31.6 & 19.7 & 54.7 & 65.3 & 51.5 & 40.1 \\
539M & 500M & 25.1 & 35.3 & 34.3 & 23.6 & 57.4 & 64.5 & 52.5 & 41.8 \\
539M & 1B   & 25.2 & 37.3 & 35.1 & 24.3 & 59.3 & 65.2 & 51.4 & \textbf{42.6} \\
\bottomrule
\end{tabular}
\end{table}

\begin{table}[h]
\centering
\caption{English downstream benchmark accuracy (\%) at the same checkpoints. More target data (higher $h$) reduces English performance, reflecting the trade-off between target and non-target domains.}
\label{tab:downstream_en}
\begin{tabular}{ll ccccccc c}
\toprule
Model & Pool & ARC-C & ARC-E & HSwag & Lambada & PIQA & SciQ & Wino & \textbf{Avg} \\
\midrule
101M & 50M  & 23.1 & 40.2 & 32.9 & 31.4 & 64.7 & 62.7 & 50.7 & 43.7 \\
101M & 100M & 23.1 & 40.2 & 32.6 & 30.8 & 64.9 & 65.3 & 49.8 & 43.8 \\
101M & 500M & 21.3 & 35.3 & 28.8 & 24.6 & 61.6 & 57.8 & 51.5 & 40.1 \\
101M & 1B   & 21.8 & 35.5 & 28.0 & 22.5 & 60.7 & 57.8 & 51.5 & 39.7 \\
\midrule
143M & 50M  & 23.2 & 40.4 & 34.6 & 36.6 & 65.2 & 64.0 & 50.5 & 44.9 \\
143M & 100M & 24.6 & 41.2 & 34.9 & 32.5 & 66.5 & 67.7 & 51.0 & 45.5 \\
143M & 500M & 23.3 & 38.8 & 32.2 & 29.9 & 64.6 & 66.0 & 48.9 & 43.4 \\
143M & 1B   & 22.4 & 38.9 & 30.3 & 28.8 & 63.0 & 62.8 & 51.9 & 42.6 \\
\midrule
192M & 50M  & 24.7 & 41.3 & 37.9 & 35.5 & 67.3 & 66.6 & 51.9 & 46.4 \\
192M & 100M & 23.9 & 41.3 & 36.8 & 35.8 & 67.7 & 67.4 & 51.2 & 46.3 \\
192M & 500M & 23.4 & 41.0 & 35.0 & 32.0 & 65.8 & 66.1 & 50.7 & 44.9 \\
192M & 1B   & 22.1 & 38.6 & 31.3 & 29.4 & 63.0 & 62.7 & 51.8 & 42.7 \\
\midrule
340M & 50M  & 24.2 & 45.2 & 41.9 & 39.8 & 68.9 & 69.7 & 52.9 & 48.9 \\
340M & 100M & 24.3 & 44.6 & 42.3 & 40.6 & 69.2 & 69.9 & 51.2 & 48.9 \\
340M & 500M & 25.4 & 44.2 & 40.1 & 40.0 & 68.5 & 70.1 & 51.7 & 48.6 \\
340M & 1B   & 24.3 & 42.3 & 38.5 & 38.5 & 66.2 & 67.6 & 51.3 & 47.0 \\
\midrule
539M & 50M  & 26.4 & 46.6 & 46.5 & 45.4 & 70.3 & 71.4 & 52.8 & \textbf{51.3} \\
539M & 100M & 24.7 & 47.5 & 46.5 & 45.4 & 69.5 & 72.4 & 53.7 & 51.4 \\
539M & 500M & 24.6 & 46.6 & 45.2 & 43.1 & 69.2 & 70.1 & 52.4 & 50.2 \\
539M & 1B   & 24.3 & 45.4 & 42.3 & 42.1 & 67.5 & 69.0 & 49.7 & 48.6 \\
\bottomrule
\end{tabular}
\end{table}

\newpage
\section{Limitations}
\label{sec:limitations}
We did not tune training hyperparameters jointly with the mixture and repetition settings, instead using standard values throughout. It is possible that interactions between learning rate, batch size, or schedule and the number of repetitions could shift the optimal operating points we report.
Additionally, all experiments use a single architecture family (GPT-2-style decoder-only transformers); whether the observed scaling behavior transfers to other architectures remains an open question. We leave both explorations to future work.
Our largest model is 939M parameters; while we observe consistent trends across model sizes suggesting the scaling law extrapolates, verifying this at scales of tens or hundreds of billions of parameters would require substantially more compute than available for this study.

%% file: main.bbl
\begin{thebibliography}{55}
\providecommand{\natexlab}[1]{#1}
\providecommand{\url}[1]{\texttt{#1}}
\expandafter\ifx\csname urlstyle\endcsname\relax
  \providecommand{\doi}[1]{doi: #1}\else
  \providecommand{\doi}{doi: \begingroup \urlstyle{rm}\Url}\fi

\bibitem[Abnar et~al.(2025)Abnar, Shah, Busbridge, El-Nouby, Susskind, and Thilak]{abnar2025parameters}
Samira Abnar, Harshay Shah, Dan Busbridge, Alaaeldin El-Nouby, Joshua~M Susskind, and Vimal Thilak.
\newblock Parameters vs flops: Scaling laws for optimal sparsity for mixture-of-experts language models.
\newblock In \emph{International Conference on Machine Learning}, pages 204--230. PMLR, 2025.

\bibitem[Anonymous(2026)]{anonymous2026mixdonttune}
Anonymous.
\newblock Mix, don't tune: Bilingual pre-training outperforms hyperparameter search in data-constrained settings.
\newblock \emph{Submitted to NeurIPS 2026}, 2026.

\bibitem[Anthropic(2026)]{antropic2026}
Anthropic.
\newblock Introducing claude opus 4.6.
\newblock \emph{Antropic Annoucements}, 2026.
\newblock URL \url{https://www.anthropic.com/news/claude-opus-4-6}.

\bibitem[Bakouch et~al.(2025)Bakouch, Ben~Allal, Lozhkov, Tazi, Tunstall, Patiño, Beeching, Roucher, Reedi, Gallouédec, Rasul, Habib, Fourrier, Kydlicek, Penedo, Larcher, Morlon, Srivastav, Lochner, Nguyen, Raffel, von Werra, and Wolf]{bakouch2025smollm3}
Elie Bakouch, Loubna Ben~Allal, Anton Lozhkov, Nouamane Tazi, Lewis Tunstall, Carlos~Miguel Patiño, Edward Beeching, Aymeric Roucher, Aksel~Joonas Reedi, Quentin Gallouédec, Kashif Rasul, Nathan Habib, Clémentine Fourrier, Hynek Kydlicek, Guilherme Penedo, Hugo Larcher, Mathieu Morlon, Vaibhav Srivastav, Joshua Lochner, Xuan-Son Nguyen, Colin Raffel, Leandro von Werra, and Thomas Wolf.
\newblock {SmolLM3: smol, multilingual, long-context reasoner}.
\newblock \url{https://huggingface.co/blog/smollm3}, 2025.

\bibitem[B{\'e}thune et~al.(2025)B{\'e}thune, Grangier, Busbridge, Gualdoni, Cuturi, and Ablin]{bethune2025scaling}
Louis B{\'e}thune, David Grangier, Dan Busbridge, Eleonora Gualdoni, Marco Cuturi, and Pierre Ablin.
\newblock Scaling laws for forgetting during finetuning with pretraining data injection.
\newblock In \emph{Forty-second International Conference on Machine Learning}, 2025.
\newblock URL \url{https://openreview.net/forum?id=vWMij23BmQ}.

\bibitem[Bisk et~al.(2020)Bisk, Zellers, Bras, Gao, and Choi]{bisk2020piqa}
Yonatan Bisk, Rowan Zellers, Ronan~Le Bras, Jianfeng Gao, and Yejin Choi.
\newblock Piqa: Reasoning about physical commonsense in natural language.
\newblock In \emph{Thirty-Fourth AAAI Conference on Artificial Intelligence}, 2020.

\bibitem[Chang et~al.(2024)Chang, Paltenghi, Li, Lin, Zhao, Huber, Liu, Rabatin, Shi, and Chandra]{chang2024scaling}
Ernie Chang, Matteo Paltenghi, Yang Li, Pin-Jie Lin, Changsheng Zhao, Patrick Huber, Zechun Liu, Rastislav Rabatin, Yangyang Shi, and Vikas Chandra.
\newblock Scaling parameter-constrained language models with quality data.
\newblock In Franck Dernoncourt, Daniel Preo{\c{t}}iuc-Pietro, and Anastasia Shimorina, editors, \emph{Proceedings of the 2024 Conference on Empirical Methods in Natural Language Processing: Industry Track}, pages 80--97, Miami, Florida, US, November 2024. Association for Computational Linguistics.
\newblock \doi{10.18653/v1/2024.emnlp-industry.8}.
\newblock URL \url{https://aclanthology.org/2024.emnlp-industry.8/}.

\bibitem[Clark et~al.(2018)Clark, Cowhey, Etzioni, Khot, Sabharwal, Schoenick, and Tafjord]{clark2018think}
Peter Clark, Isaac Cowhey, Oren Etzioni, Tushar Khot, Ashish Sabharwal, Carissa Schoenick, and Oyvind Tafjord.
\newblock Think you have solved question answering? try arc, the ai2 reasoning challenge.
\newblock \emph{arXiv preprint arXiv:1803.05457}, 2018.

\bibitem[Diao et~al.(2025)Diao, Yang, Fu, Dong, Su, Kliegl, Chen, Belcak, Suhara, Yin, Patwary, Yingyan, Lin, Kautz, and Molchanov]{diaonemotron}
Shizhe Diao, Yu~Yang, Yonggan Fu, Xin Dong, Dan Su, Markus Kliegl, Zijia Chen, Peter Belcak, Yoshi Suhara, Hongxu Yin, Mostofa Patwary, Yingyan, Lin, Jan Kautz, and Pavlo Molchanov.
\newblock Nemotron-climb: Clustering-based iterative data mixture bootstrapping for language model pre-training, 2025.
\newblock URL \url{https://arxiv.org/abs/2504.13161}.

\bibitem[{Essential AI} et~al.(2025){Essential AI}, :, Hojel, Pust, Romanski, Vanjani, Kapila, Parmar, Chaluvaraju, Tripathy, Thomas, Tanwer, Shah, Shah, Stratos, Nguyen, Smith, Callahan, Rushton, Monk, Mazarakis, Jamal, Srivastava, Singla, and Vaswani]{hojel2025essential}
{Essential AI}, :, Andrew Hojel, Michael Pust, Tim Romanski, Yash Vanjani, Ritvik Kapila, Mohit Parmar, Adarsh Chaluvaraju, Alok Tripathy, Anil Thomas, Ashish Tanwer, Darsh~J Shah, Ishaan Shah, Karl Stratos, Khoi Nguyen, Kurt Smith, Michael Callahan, Peter Rushton, Philip Monk, Platon Mazarakis, Saad Jamal, Saurabh Srivastava, Somanshu Singla, and Ashish Vaswani.
\newblock Essential-web v1.0: 24t tokens of organized web data, 2025.
\newblock URL \url{https://arxiv.org/abs/2506.14111}.

\bibitem[Fan et~al.(2024)Fan, Pagliardini, and Jaggi]{fan2024doge}
Simin Fan, Matteo Pagliardini, and Martin Jaggi.
\newblock Doge: domain reweighting with generalization estimation.
\newblock In \emph{Proceedings of the 41st International Conference on Machine Learning}, ICML'24. JMLR.org, 2024.

\bibitem[Gao et~al.(2020)Gao, Biderman, Black, Golding, Hoppe, Foster, Phang, He, Thite, Nabeshima, Presser, and Leahy]{gao2020pile}
Leo Gao, Stella Biderman, Sid Black, Laurence Golding, Travis Hoppe, Charles Foster, Jason Phang, Horace He, Anish Thite, Noa Nabeshima, Shawn Presser, and Connor Leahy.
\newblock The {P}ile: An 800gb dataset of diverse text for language modeling.
\newblock \emph{arXiv preprint arXiv:2101.00027}, 2020.

\bibitem[Goyal et~al.(2024)Goyal, Maini, Lipton, Raghunathan, and Kolter]{goyal2024scaling}
Sachin Goyal, Pratyush Maini, Zachary~C. Lipton, Aditi Raghunathan, and J.~Zico Kolter.
\newblock Scaling laws for data filtering—data curation cannot be compute agnostic.
\newblock In \emph{2024 IEEE/CVF Conference on Computer Vision and Pattern Recognition (CVPR)}, pages 22702--22711, 2024.
\newblock \doi{10.1109/CVPR52733.2024.02142}.

\bibitem[Grangier et~al.(2025)Grangier, Fan, Seto, and Ablin]{grangier2025task}
David Grangier, Simin Fan, Skyler Seto, and Pierre Ablin.
\newblock Task-adaptive pretrained language models via clustered-importance sampling.
\newblock In \emph{ICLR}, 2025.

\bibitem[Gunasekar et~al.(2023)Gunasekar, Zhang, Aneja, Cesar, Mendes, Giorno, Gopi, Javaheripi, Kauffmann, de~Rosa, Saarikivi, Salim, Shah, Singh~Behl, Wang, Bubeck, Eldan, Kalai, Lee, and Li]{gunasekar2023textbooks}
Suriya Gunasekar, Yi~Zhang, Jyoti Aneja, Caio Cesar, Teodoro Mendes, Allie~Del Giorno, Sivakanth Gopi, Mojan Javaheripi, Piero Kauffmann, Gustavo de~Rosa, Olli Saarikivi, Adil Salim, Shital Shah, Harkirat Singh~Behl, Xin Wang, Sébastien Bubeck, Ronen Eldan, Adam~Tauman Kalai, Yin~Tat Lee, and Yuanzhi Li.
\newblock Textbooks are all you need, June 2023.
\newblock URL \url{https://www.microsoft.com/en-us/research/publication/textbooks-are-all-you-need/}.

\bibitem[H{\"a}gele et~al.(2024)H{\"a}gele, Bakouch, Kosson, allal, Werra, and Jaggi]{hagele2024scaling}
Alexander H{\"a}gele, Elie Bakouch, Atli Kosson, Loubna~Ben allal, Leandro~Von Werra, and Martin Jaggi.
\newblock Scaling laws and compute-optimal training beyond fixed training durations.
\newblock In \emph{The Thirty-eighth Annual Conference on Neural Information Processing Systems}, 2024.
\newblock URL \url{https://openreview.net/forum?id=Y13gSfTjGr}.

\bibitem[Hernandez et~al.(2022)Hernandez, Brown, Conerly, DasSarma, Drain, El-Showk, Elhage, Hatfield-Dodds, Henighan, Hume, Johnston, Mann, Olah, Olsson, Amodei, Joseph, Kaplan, and McCandlish]{hernandez2022scaling}
Danny Hernandez, Tom Brown, Tom Conerly, Nova DasSarma, Dawn Drain, Sheer El-Showk, Nelson Elhage, Zac Hatfield-Dodds, Tom Henighan, Tristan Hume, Scott Johnston, Ben Mann, Chris Olah, Catherine Olsson, Dario Amodei, Nicholas Joseph, Jared Kaplan, and Sam McCandlish.
\newblock Scaling laws and interpretability of learning from repeated data, 2022.
\newblock URL \url{https://arxiv.org/abs/2205.10487}.

\bibitem[Hoffmann et~al.(2022)Hoffmann, Borgeaud, Mensch, Buchatskaya, Cai, Rutherford, de~Las~Casas, Hendricks, Welbl, Clark, Hennigan, Noland, Millican, van~den Driessche, Damoc, Guy, Osindero, Simonyan, Elsen, Vinyals, Rae, and Sifre]{hoffmann2022training}
Jordan Hoffmann, Sebastian Borgeaud, Arthur Mensch, Elena Buchatskaya, Trevor Cai, Eliza Rutherford, Diego de~Las~Casas, Lisa~Anne Hendricks, Johannes Welbl, Aidan Clark, Tom Hennigan, Eric Noland, Katie Millican, George van~den Driessche, Bogdan Damoc, Aurelia Guy, Simon Osindero, Karen Simonyan, Erich Elsen, Oriol Vinyals, Jack~W. Rae, and Laurent Sifre.
\newblock Training compute-optimal large language models.
\newblock In \emph{Proceedings of the 36th International Conference on Neural Information Processing Systems}, NIPS '22, Red Hook, NY, USA, 2022. Curran Associates Inc.S.
\newblock ISBN 9781713871088.

\bibitem[Joshi et~al.(2020)Joshi, Santy, Budhiraja, Bali, and Choudhury]{joshi2020state}
Pratik Joshi, Sebastin Santy, Amar Budhiraja, Kalika Bali, and Monojit Choudhury.
\newblock The state and fate of linguistic diversity and inclusion in the {NLP} world.
\newblock In Dan Jurafsky, Joyce Chai, Natalie Schluter, and Joel Tetreault, editors, \emph{Proceedings of the 58th Annual Meeting of the Association for Computational Linguistics}, pages 6282--6293, Online, July 2020. Association for Computational Linguistics.
\newblock \doi{10.18653/v1/2020.acl-main.560}.
\newblock URL \url{https://aclanthology.org/2020.acl-main.560/}.

\bibitem[Kaplan et~al.(2020)Kaplan, McCandlish, Henighan, Brown, Chess, Child, Gray, Radford, Wu, and Amodei]{kaplan2020scaling}
Jared Kaplan, Sam McCandlish, Tom Henighan, Tom~B Brown, Benjamin Chess, Rewon Child, Scott Gray, Alec Radford, Jeffrey Wu, and Dario Amodei.
\newblock Scaling laws for neural language models.
\newblock \emph{arXiv preprint arXiv:2001.08361}, 2020.

\bibitem[Kingma and Ba(2014)]{kingma2014adam}
Diederik~P Kingma and Jimmy Ba.
\newblock Adam: A method for stochastic optimization.
\newblock \emph{arXiv preprint arXiv:1412.6980}, 2014.

\bibitem[Krajewski et~al.(2024)Krajewski, Ludziejewski, Adamczewski, Pióro, Krutul, Antoniak, Ciebiera, Król, Odrzygóźdź, Sankowski, Cygan, and Jaszczur]{krajewski2024scaling}
Jakub Krajewski, Jan Ludziejewski, Kamil Adamczewski, Maciej Pióro, Michał Krutul, Szymon Antoniak, Kamil Ciebiera, Krystian Król, Tomasz Odrzygóźdź, Piotr Sankowski, Marek Cygan, and Sebastian Jaszczur.
\newblock Scaling laws for fine-grained mixture of experts, 2024.
\newblock URL \url{https://arxiv.org/abs/2402.07871}.

\bibitem[Kudo and Richardson(2018)]{kudo2018sentencepiece}
Taku Kudo and John Richardson.
\newblock Sentencepiece: A simple and language independent subword tokenizer and detokenizer for neural text processing.
\newblock In \emph{Proceedings of the 2018 Conference on Empirical Methods in Natural Language Processing: System Demonstrations}, pages 66--71, 2018.

\bibitem[Li et~al.(2025)Li, Fang, Smyrnis, Ivgi, Jordan, Gadre, Bansal, Guha, Keh, Arora, Garg, Xin, Muennighoff, Heckel, Mercat, Chen, Gururangan, Wortsman, Albalak, Bitton, Nezhurina, Abbas, Hsieh, Ghosh, Gardner, Kilian, Zhang, Shao, Pratt, Sanyal, Ilharco, Daras, Marathe, Gokaslan, Zhang, Chandu, Nguyen, Vasiljevic, Kakade, Song, Sanghavi, Faghri, Oh, Zettlemoyer, Lo, El-Nouby, Pouransari, Toshev, Wang, Groeneveld, Soldaini, Koh, Jitsev, Kollar, Dimakis, Carmon, Dave, Schmidt, and Shankar]{li2024datacomp}
Jeffrey Li, Alex Fang, Georgios Smyrnis, Maor Ivgi, Matt Jordan, Samir Gadre, Hritik Bansal, Etash Guha, Sedrick Keh, Kushal Arora, Saurabh Garg, Rui Xin, Niklas Muennighoff, Reinhard Heckel, Jean Mercat, Mayee Chen, Suchin Gururangan, Mitchell Wortsman, Alon Albalak, Yonatan Bitton, Marianna Nezhurina, Amro Abbas, Cheng-Yu Hsieh, Dhruba Ghosh, Josh Gardner, Maciej Kilian, Hanlin Zhang, Rulin Shao, Sarah Pratt, Sunny Sanyal, Gabriel Ilharco, Giannis Daras, Kalyani Marathe, Aaron Gokaslan, Jieyu Zhang, Khyathi Chandu, Thao Nguyen, Igor Vasiljevic, Sham Kakade, Shuran Song, Sujay Sanghavi, Fartash Faghri, Sewoong Oh, Luke Zettlemoyer, Kyle Lo, Alaaeldin El-Nouby, Hadi Pouransari, Alexander Toshev, Stephanie Wang, Dirk Groeneveld, Luca Soldaini, Pang~Wei Koh, Jenia Jitsev, Thomas Kollar, Alexandros~G. Dimakis, Yair Carmon, Achal Dave, Ludwig Schmidt, and Vaishaal Shankar.
\newblock Datacomp-lm: In search of the next generation of training sets for language models, 2025.
\newblock URL \url{https://arxiv.org/abs/2406.11794}.

\bibitem[Liew and Kato(2025)]{liew2025acceleration}
Seng~Pei Liew and Takuya Kato.
\newblock From acceleration to saturation: Scaling behavior of bootstrapped language model pretraining.
\newblock In \emph{NeurIPS 2025 Workshop on Evaluating the Evolving LLM Lifecycle: Benchmarks, Emergent Abilities, and Scaling}, 2025.
\newblock URL \url{https://openreview.net/forum?id=PhsneSYvWK}.

\bibitem[Luong and Lockhart(2025)]{luong2025advanced}
Thang Luong and Edward Lockhart.
\newblock Advanced version of gemini with deep think officially achieves gold-medal standard at the international mathematical olympiad.
\newblock \emph{Google DeepMind Blog}, 1, 2025.

\bibitem[Maini et~al.(2024)Maini, Seto, Bai, Grangier, Zhang, and Jaitly]{maini2024rephrasing}
Pratyush Maini, Skyler Seto, Richard Bai, David Grangier, Yizhe Zhang, and Navdeep Jaitly.
\newblock Rephrasing the web: A recipe for compute and data-efficient language modeling.
\newblock In \emph{Proceedings of the 62nd Annual Meeting of the Association for Computational Linguistics (Volume 1: Long Papers)}, pages 14044--14072, 2024.

\bibitem[Muennighoff et~al.(2023)Muennighoff, Rush, Barak, Scao, Tazi, Piktus, Pyysalo, Wolf, and Raffel]{muennighoff2023scaling}
Niklas Muennighoff, Alexander~M Rush, Boaz Barak, Teven~Le Scao, Nouamane Tazi, Aleksandra Piktus, Sampo Pyysalo, Thomas Wolf, and Colin Raffel.
\newblock Scaling data-constrained language models.
\newblock In \emph{Thirty-seventh Conference on Neural Information Processing Systems}, 2023.
\newblock URL \url{https://openreview.net/forum?id=j5BuTrEj35}.

\bibitem[OLMo et~al.(2025{\natexlab{a}})OLMo, :, Ettinger, Bertsch, Kuehl, Graham, Heineman, Groeneveld, Brahman, Timbers, Ivison, Morrison, Poznanski, Lo, Soldaini, Jordan, Chen, Noukhovitch, Lambert, Walsh, Dasigi, Berry, Malik, Shah, Geng, Arora, Gupta, Anderson, Xiao, Murray, Romero, Graf, Asai, Bhagia, Wettig, Liu, Rangapur, Anastasiades, Huang, Schwenk, Trivedi, Magnusson, Lochner, Liu, Miranda, Sap, Morgan, Schmitz, Guerquin, Wilson, Huff, Bras, Xin, Shao, Skjonsberg, Shen, Li, Wilde, Pyatkin, Merrill, Chang, Gu, Zeng, Sabharwal, Zettlemoyer, Koh, Farhadi, Smith, and Hajishirzi]{olmo2025olmo}
Team OLMo, :, Allyson Ettinger, Amanda Bertsch, Bailey Kuehl, David Graham, David Heineman, Dirk Groeneveld, Faeze Brahman, Finbarr Timbers, Hamish Ivison, Jacob Morrison, Jake Poznanski, Kyle Lo, Luca Soldaini, Matt Jordan, Mayee Chen, Michael Noukhovitch, Nathan Lambert, Pete Walsh, Pradeep Dasigi, Robert Berry, Saumya Malik, Saurabh Shah, Scott Geng, Shane Arora, Shashank Gupta, Taira Anderson, Teng Xiao, Tyler Murray, Tyler Romero, Victoria Graf, Akari Asai, Akshita Bhagia, Alexander Wettig, Alisa Liu, Aman Rangapur, Chloe Anastasiades, Costa Huang, Dustin Schwenk, Harsh Trivedi, Ian Magnusson, Jaron Lochner, Jiacheng Liu, Lester James~V. Miranda, Maarten Sap, Malia Morgan, Michael Schmitz, Michal Guerquin, Michael Wilson, Regan Huff, Ronan~Le Bras, Rui Xin, Rulin Shao, Sam Skjonsberg, Shannon~Zejiang Shen, Shuyue~Stella Li, Tucker Wilde, Valentina Pyatkin, Will Merrill, Yapei Chang, Yuling Gu, Zhiyuan Zeng, Ashish Sabharwal, Luke Zettlemoyer, Pang~Wei Koh, Ali Farhadi, Noah~A. Smith, and Hannaneh
  Hajishirzi.
\newblock Olmo 3, 2025{\natexlab{a}}.
\newblock URL \url{https://arxiv.org/abs/2512.13961}.

\bibitem[OLMo et~al.(2025{\natexlab{b}})OLMo, Walsh, Soldaini, Groeneveld, Lo, Arora, Bhagia, Gu, Huang, Jordan, Lambert, Schwenk, Tafjord, Anderson, Atkinson, Brahman, Clark, Dasigi, Dziri, Ettinger, Guerquin, Heineman, Ivison, Koh, Liu, Malik, Merrill, Miranda, Morrison, Murray, Nam, Poznanski, Pyatkin, Rangapur, Schmitz, Skjonsberg, Wadden, Wilhelm, Wilson, Zettlemoyer, Farhadi, Smith, and Hajishirzi]{walsh2024olmo2}
Team OLMo, Pete Walsh, Luca Soldaini, Dirk Groeneveld, Kyle Lo, Shane Arora, Akshita Bhagia, Yuling Gu, Shengyi Huang, Matt Jordan, Nathan Lambert, Dustin Schwenk, Oyvind Tafjord, Taira Anderson, David Atkinson, Faeze Brahman, Christopher Clark, Pradeep Dasigi, Nouha Dziri, Allyson Ettinger, Michal Guerquin, David Heineman, Hamish Ivison, Pang~Wei Koh, Jiacheng Liu, Saumya Malik, William Merrill, Lester James~V. Miranda, Jacob Morrison, Tyler Murray, Crystal Nam, Jake Poznanski, Valentina Pyatkin, Aman Rangapur, Michael Schmitz, Sam Skjonsberg, David Wadden, Christopher Wilhelm, Michael Wilson, Luke Zettlemoyer, Ali Farhadi, Noah~A. Smith, and Hannaneh Hajishirzi.
\newblock 2 olmo 2 furious, 2025{\natexlab{b}}.
\newblock URL \url{https://arxiv.org/abs/2501.00656}.

\bibitem[Paperno et~al.(2016)Paperno, Kruszewski, Lazaridou, Pham, Bernardi, Pezzelle, Baroni, Boleda, and Fern{\'a}ndez]{paperno2016lambada}
Denis Paperno, Germ{\'a}n Kruszewski, Angeliki Lazaridou, Ngoc~Quan Pham, Raffaella Bernardi, Sandro Pezzelle, Marco Baroni, Gemma Boleda, and Raquel Fern{\'a}ndez.
\newblock The {LAMBADA} dataset: Word prediction requiring a broad discourse context.
\newblock In Katrin Erk and Noah~A. Smith, editors, \emph{Proceedings of the 54th Annual Meeting of the Association for Computational Linguistics (Volume 1: Long Papers)}, pages 1525--1534, Berlin, Germany, August 2016. Association for Computational Linguistics.
\newblock \doi{10.18653/v1/P16-1144}.
\newblock URL \url{https://aclanthology.org/P16-1144/}.

\bibitem[Paster et~al.(2023)Paster, Santos, Azerbayev, and Ba]{paster2023openwebmath}
Keiran Paster, Marco~Dos Santos, Zhangir Azerbayev, and Jimmy Ba.
\newblock {OpenWebMath}: An open dataset of high-quality mathematical web text.
\newblock \emph{arXiv preprint arXiv:2310.06786}, 2023.

\bibitem[Penedo et~al.(2024)Penedo, Kydl{\'\i}{\v{c}}ek, Ben~allal, Lozhkov, Mitchell, Raffel, Von~Werra, and Wolf]{penedo2024fineweb}
Guilherme Penedo, Hynek Kydl{\'\i}{\v{c}}ek, Loubna Ben~allal, Anton Lozhkov, Margaret Mitchell, Colin Raffel, Leandro Von~Werra, and Thomas Wolf.
\newblock The {FineWeb} datasets: Decanting the web for the finest text data at scale.
\newblock In \emph{Advances in Neural Information Processing Systems}, volume~37, 2024.

\bibitem[Penedo et~al.(2025)Penedo, Kydl{\'\i}{\v{c}}ek, Sabolčec, Messmer, Foroutan, Kargaran, Raffel, Jaggi, Von~Werra, and Wolf]{penedo2025fineweb2}
Guilherme Penedo, Hynek Kydl{\'\i}{\v{c}}ek, Vinko Sabolčec, Bettina Messmer, Negar Foroutan, Amir~Hossein Kargaran, Colin Raffel, Martin Jaggi, Leandro Von~Werra, and Thomas Wolf.
\newblock {FineWeb2}: One pipeline to scale them all -- adapting pre-training data processing to every language.
\newblock \emph{arXiv preprint arXiv:2506.20920}, 2025.

\bibitem[Porian et~al.(2024)Porian, Wortsman, Jitsev, Schmidt, and Carmon]{porian2025resolving}
Tomer Porian, Mitchell Wortsman, Jenia Jitsev, Ludwig Schmidt, and Yair Carmon.
\newblock Resolving discrepancies in compute-optimal scaling of language models.
\newblock In \emph{The Thirty-eighth Annual Conference on Neural Information Processing Systems}, 2024.
\newblock URL \url{https://openreview.net/forum?id=4fSSqpk1sM}.

\bibitem[Que et~al.(2024)Que, Liu, Zhang, Zhang, Qu, Ma, Duan, ZhiqiBai, JiakaiWang, Zhang, Tan, Fu, Wang, Qu, Su, and Zheng]{que2024d}
Haoran Que, Jiaheng Liu, Ge~Zhang, Chenchen Zhang, Xingwei Qu, Yinghao Ma, Feiyu Duan, ZhiqiBai, JiakaiWang, Yuanxing Zhang, Xu~Tan, Jie Fu, Jiamang Wang, Lin Qu, Wenbo Su, and Bo~Zheng.
\newblock D-{CPT} law: Domain-specific continual pre-training scaling law for large language models.
\newblock In \emph{The Thirty-eighth Annual Conference on Neural Information Processing Systems}, 2024.
\newblock URL \url{https://openreview.net/forum?id=JzKFN5fWOk}.

\bibitem[Radford et~al.(2019)Radford, Wu, Child, Luan, Amodei, and Sutskever]{radford2019language}
Alec Radford, Jeffrey Wu, Rewon Child, David Luan, Dario Amodei, and Ilya Sutskever.
\newblock Language models are unsupervised multitask learners.
\newblock \emph{OpenAI blog}, 2019.

\bibitem[Raffel et~al.(2020)Raffel, Shazeer, Roberts, Lee, Narang, Matena, Zhou, Li, and Liu]{raffel2020exploring}
Colin Raffel, Noam Shazeer, Adam Roberts, Katherine Lee, Sharan Narang, Michael Matena, Yanqi Zhou, Wei Li, and Peter~J Liu.
\newblock Exploring the limits of transfer learning with a unified text-to-text transformer.
\newblock \emph{Journal of Machine Learning Research}, 21\penalty0 (140):\penalty0 1--67, 2020.

\bibitem[Sakaguchi et~al.(2020)Sakaguchi, Le~Bras, Bhagavatula, and Choi]{sakaguchi2020winogrande}
Keisuke Sakaguchi, Ronan Le~Bras, Chandra Bhagavatula, and Yejin Choi.
\newblock Winogrande: An adversarial winograd schema challenge at scale.
\newblock \emph{Proceedings of AAAI}, 2020.

\bibitem[Seto et~al.(2025)Seto, Ter~Hoeve, Bai, Schluter, and Grangier]{seto2025training}
Skyler Seto, Maartje Ter~Hoeve, Richard~He Bai, Natalie Schluter, and David Grangier.
\newblock Training bilingual lms with data constraints in the targeted language.
\newblock In \emph{Findings of the Association for Computational Linguistics: ACL 2025}, pages 19096--19122, 2025.

\bibitem[Seto et~al.(2026)Seto, Ablin, Filippova, Ye, Bethune, Katharopoulos, and Grangier]{seto2026optimal}
Skyler Seto, Pierre Ablin, Anastasiia Filippova, Jiayuan Ye, Louis Bethune, Angelos Katharopoulos, and David Grangier.
\newblock Optimal splitting of language models from mixtures to specialized domains.
\newblock \emph{arXiv preprint arXiv:2603.19149}, 2026.

\bibitem[Shukor et~al.(2025)Shukor, Bethune, Busbridge, Grangier, Fini, El-Nouby, and Ablin]{shukor2025scalinglawsoptimaldata}
Mustafa Shukor, Louis Bethune, Dan Busbridge, David Grangier, Enrico Fini, Alaaeldin El-Nouby, and Pierre Ablin.
\newblock Scaling laws for optimal data mixtures.
\newblock In \emph{NeurIPS}, 2025.
\newblock URL \url{https://arxiv.org/abs/2507.09404}.

\bibitem[Singh et~al.(2026)Singh, Fry, Perelman, Tart, Ganesh, El-Kishky, McLaughlin, Low, Ostrow, Ananthram, et~al.]{singh2025openai}
Aaditya Singh, Adam Fry, Adam Perelman, Adam Tart, Adi Ganesh, Ahmed El-Kishky, Aidan McLaughlin, Aiden Low, AJ~Ostrow, Akhila Ananthram, et~al.
\newblock Openai gpt-5 system card.
\newblock \emph{arXiv preprint arXiv:2601.03267}, 2026.

\bibitem[Soboleva et~al.(2023)Soboleva, Al-Khateeb, Myers, Steeves, Hestness, and Dey]{cerebras2023slimpajama}
Daria Soboleva, Faisal Al-Khateeb, Robert Myers, Jacob~R Steeves, Joel Hestness, and Nolan Dey.
\newblock {SlimPajama: A 627B token cleaned and deduplicated version of RedPajama}.
\newblock \url{https://cerebras.ai/blog/slimpajama-a-627b-token-cleaned-and-deduplicated-version-of-redpajama}, 2023.
\newblock URL \url{https://huggingface.co/datasets/cerebras/SlimPajama-627B}.

\bibitem[Soldaini and Lo(2023)]{soldaini2023pes2o}
Luca Soldaini and Kyle Lo.
\newblock {peS2o} (pretraining efficiently on {S2ORC}) dataset.
\newblock Technical report, Allen Institute for AI, 2023.

\bibitem[Soldaini et~al.(2024)Soldaini, Kinney, Bhagia, Schwenk, Atkinson, Authur, Bogin, Chandu, Dumas, Elazar, Hofmann, Jha, Kumar, Lucy, Lyu, Lambert, Magnusson, Morrison, Muennighoff, Naik, Nam, Peters, Ravichander, Richardson, Shen, Strubell, Subramani, Tafjord, Walsh, Zettlemoyer, Smith, Hajishirzi, Beltagy, Groeneveld, Dodge, and Lo]{soldaini2024dolma}
Luca Soldaini, Rodney Kinney, Akshita Bhagia, Dustin Schwenk, David Atkinson, Russell Authur, Ben Bogin, Khyathi Chandu, Jennifer Dumas, Yanai Elazar, Valentin Hofmann, Ananya Jha, Sachin Kumar, Li~Lucy, Xinxi Lyu, Nathan Lambert, Ian Magnusson, Jacob Morrison, Niklas Muennighoff, Aakanksha Naik, Crystal Nam, Matthew Peters, Abhilasha Ravichander, Kyle Richardson, Zejiang Shen, Emma Strubell, Nishant Subramani, Oyvind Tafjord, Evan Walsh, Luke Zettlemoyer, Noah Smith, Hannaneh Hajishirzi, Iz~Beltagy, Dirk Groeneveld, Jesse Dodge, and Kyle Lo.
\newblock Dolma: an open corpus of three trillion tokens for language model pretraining research.
\newblock In Lun-Wei Ku, Andre Martins, and Vivek Srikumar, editors, \emph{Proceedings of the 62nd Annual Meeting of the Association for Computational Linguistics (Volume 1: Long Papers)}, pages 15725--15788, Bangkok, Thailand, August 2024. Association for Computational Linguistics.
\newblock \doi{10.18653/v1/2024.acl-long.840}.
\newblock URL \url{https://aclanthology.org/2024.acl-long.840/}.

\bibitem[Su et~al.(2025)Su, Kong, Lin, Jennings, Norick, Kliegl, Patwary, Shoeybi, and Catanzaro]{dash2024nemotroncc}
Dan Su, Kezhi Kong, Ying Lin, Joseph Jennings, Brandon Norick, Markus Kliegl, Mostofa Patwary, Mohammad Shoeybi, and Bryan Catanzaro.
\newblock Nemotron-{CC}: Transforming {C}ommon {C}rawl into a refined long-horizon pretraining dataset.
\newblock In Wanxiang Che, Joyce Nabende, Ekaterina Shutova, and Mohammad~Taher Pilehvar, editors, \emph{Proceedings of the 63rd Annual Meeting of the Association for Computational Linguistics (Volume 1: Long Papers)}, pages 2459--2475, Vienna, Austria, July 2025. Association for Computational Linguistics.
\newblock ISBN 979-8-89176-251-0.
\newblock \doi{10.18653/v1/2025.acl-long.123}.
\newblock URL \url{https://aclanthology.org/2025.acl-long.123/}.

\bibitem[Wang et~al.(2024)Wang, Chen, Li, He, Zhang, and Wang]{wang2024scaling}
Siqi Wang, Zhengyu Chen, Bei Li, Keqing He, Min Zhang, and Jingang Wang.
\newblock Scaling laws across model architectures: A comparative analysis of dense and {M}o{E} models in large language models.
\newblock In Yaser Al-Onaizan, Mohit Bansal, and Yun-Nung Chen, editors, \emph{Proceedings of the 2024 Conference on Empirical Methods in Natural Language Processing}, pages 5583--5595, Miami, Florida, USA, November 2024. Association for Computational Linguistics.
\newblock \doi{10.18653/v1/2024.emnlp-main.319}.
\newblock URL \url{https://aclanthology.org/2024.emnlp-main.319/}.

\bibitem[Wei et~al.(2022)Wei, Tay, Bommasani, Raffel, Zoph, Borgeaud, Yogatama, Bosma, Zhou, Metzler, Chi, Hashimoto, Vinyals, Liang, Dean, and Fedus]{wei2022emergent}
Jason Wei, Yi~Tay, Rishi Bommasani, Colin Raffel, Barret Zoph, Sebastian Borgeaud, Dani Yogatama, Maarten Bosma, Denny Zhou, Donald Metzler, Ed~H. Chi, Tatsunori Hashimoto, Oriol Vinyals, Percy Liang, Jeff Dean, and William Fedus.
\newblock Emergent abilities of large language models.
\newblock \emph{Transactions on Machine Learning Research}, 2022.
\newblock ISSN 2835-8856.
\newblock URL \url{https://openreview.net/forum?id=yzkSU5zdwD}.
\newblock Survey Certification.

\bibitem[Welbl et~al.(2017)Welbl, Liu, and Gardner]{welbl2017sciq}
Johannes Welbl, Nelson~F. Liu, and Matt Gardner.
\newblock Crowdsourcing multiple choice science questions.
\newblock In Leon Derczynski, Wei Xu, Alan Ritter, and Tim Baldwin, editors, \emph{Proceedings of the 3rd Workshop on Noisy User-generated Text}, pages 94--106, Copenhagen, Denmark, September 2017. Association for Computational Linguistics.
\newblock \doi{10.18653/v1/W17-4413}.
\newblock URL \url{https://aclanthology.org/W17-4413/}.

\bibitem[Woodruff et~al.(2026)Woodruff, Cohen-Addad, Jain, Mao, Zuo, Bateni, Branzei, Brenner, Chen, Feng, Fortnow, Fu, Guan, Hadizadeh, Hajiaghayi, JafariRaviz, Javanmard, S., ichi Kawarabayashi, Kumar, Lattanzi, Lee, Li, Panageas, Paparas, Przybocki, Subercaseaux, Svensson, Taherijam, Wu, Yogev, Zadimoghaddam, Zhou, Matias, Manyika, and Mirrokni]{woodruff2026accelerating}
David~P. Woodruff, Vincent Cohen-Addad, Lalit Jain, Jieming Mao, Song Zuo, MohammadHossein Bateni, Simina Branzei, Michael~P. Brenner, Lin Chen, Ying Feng, Lance Fortnow, Gang Fu, Ziyi Guan, Zahra Hadizadeh, Mohammad~T. Hajiaghayi, Mahdi JafariRaviz, Adel Javanmard, Karthik~C. S., Ken ichi Kawarabayashi, Ravi Kumar, Silvio Lattanzi, Euiwoong Lee, Yi~Li, Ioannis Panageas, Dimitris Paparas, Benjamin Przybocki, Bernardo Subercaseaux, Ola Svensson, Shayan Taherijam, Xuan Wu, Eylon Yogev, Morteza Zadimoghaddam, Samson Zhou, Yossi Matias, James Manyika, and Vahab Mirrokni.
\newblock Accelerating scientific research with gemini: Case studies and common techniques, 2026.
\newblock URL \url{https://arxiv.org/abs/2602.03837}.

\bibitem[Xie et~al.(2023)Xie, Pham, Dong, Du, Liu, Lu, Liang, Le, Ma, and Yu]{xie2024doremi}
Sang~Michael Xie, Hieu Pham, Xuanyi Dong, Nan Du, Hanxiao Liu, Yifeng Lu, Percy Liang, Quoc~V Le, Tengyu Ma, and Adams~Wei Yu.
\newblock {DoReMi}: Optimizing data mixtures speeds up language model pretraining.
\newblock \emph{Advances in Neural Information Processing Systems}, 36, 2023.

\bibitem[Ye et~al.(2025)Ye, Liu, Sun, Zhan, Zhou, and Qiu]{ye2024data}
Jiasheng Ye, Peiju Liu, Tianxiang Sun, Jun Zhan, Yunhua Zhou, and Xipeng Qiu.
\newblock Data mixing laws: Optimizing data mixtures by predicting language modeling performance.
\newblock In \emph{The Thirteenth International Conference on Learning Representations}, 2025.
\newblock URL \url{https://openreview.net/forum?id=jjCB27TMK3}.

\bibitem[Zellers et~al.(2019)Zellers, Holtzman, Bisk, Farhadi, and Choi]{zellers2019hellaswag}
Rowan Zellers, Ari Holtzman, Yonatan Bisk, Ali Farhadi, and Yejin Choi.
\newblock {H}ella{S}wag: Can a machine really finish your sentence?
\newblock In Anna Korhonen, David Traum, and Llu{\'i}s M{\`a}rquez, editors, \emph{Proceedings of the 57th Annual Meeting of the Association for Computational Linguistics}, pages 4791--4800, Florence, Italy, July 2019. Association for Computational Linguistics.
\newblock \doi{10.18653/v1/P19-1472}.
\newblock URL \url{https://aclanthology.org/P19-1472/}.

\bibitem[Zhang et~al.(2024)Zhang, Liu, Cherry, and Firat]{zhang2024scaling}
Biao Zhang, Zhongtao Liu, Colin Cherry, and Orhan Firat.
\newblock When scaling meets {LLM} finetuning: The effect of data, model and finetuning method.
\newblock In \emph{The Twelfth International Conference on Learning Representations}, 2024.
\newblock URL \url{https://openreview.net/forum?id=5HCnKDeTws}.

\end{thebibliography}
